\documentclass[letterpaper,twocolumn,10pt]{article}
\usepackage{main}

\usepackage{paralist, tabularx}

\usepackage{bbold}
\usepackage{amsmath}
\usepackage{wrapfig}
\usepackage{amssymb}

\usepackage{lipsum}
\usepackage{hyperref}
\usepackage{url}
\usepackage{comment}
\usepackage{algorithm}
\usepackage{algpseudocode}
\usepackage{subfigmat}
\usepackage{array}
\usepackage{multirow}
\newcommand{\clientbox}[1]{\colorbox{myblue!30}{#1}}
\newcommand{\serverbox}[1]{\colorbox{mygreen!30}{#1}}

\definecolor{mygreen}{RGB}{77,175,74}
\definecolor{myblue}{RGB}{55,126,184}
\definecolor{skyblue}{RGB}{117,187,253}
\definecolor{myred}{RGB}{228,26,28}

\newcommand{\paragraphb}[1]{\noindent{\bf #1} }

\usepackage{colortbl}

\usepackage{tikz}
\newcommand*\circled[1]{\tikz[baseline=(char.base)]{
            \node[shape=circle,draw,inner sep=.3pt] (char) {#1};}}

\newcommand{\Name}[0]{FRL}

\usepackage{filecontents}

\begin{document}

\date{}

\title{FRL: Federated Rank Learning}
\author{
{\rm Hamid Mozaffari, \rm Virat Shejwalkar \& \rm Amir Houmansadr}\\
University of Massachusetts Amherst \\
\{hamid, vshejwalkar, amir\}@cs.umass.edu
} 

\maketitle

\begin{abstract}
Federated learning (FL) allows mutually untrusted clients  to collaboratively train a common machine learning model without sharing their private/proprietary training data among each other.
FL is unfortunately susceptible to \emph{poisoning} by malicious clients who aim to  hamper the accuracy of the commonly trained  model through sending malicious model updates during FL's training process.  

We argue that the key factor to the success of poisoning attacks against  existing FL systems is the large space of model updates  available to the clients, allowing malicious clients to search for the most poisonous model updates, e.g., by solving an optimization problem. 
To address this, we propose \emph{\textbf{F}ederated \textbf{R}ank \textbf{L}earning} (\Name{}). \Name{} reduces the space of client updates from model parameter updates (a continuous space of float numbers) in standard FL to the space of parameter rankings (a discrete space of integer values). To be able to train the global model using parameter ranks (instead of parameter weights), \Name{} leverage ideas from recent \emph{supermasks training} mechanisms. Specifically, \Name{} clients  rank the  parameters of a  randomly initialized neural network (provided by the server) based on their local training data. The \Name{} server uses a voting mechanism to aggregate the parameter rankings submitted by  clients in each  training epoch to generate the global ranking of the next training epoch. 

Intuitively, our voting-based aggregation mechanism prevents poisoning clients from making significant adversarial modifications to the global model, as each client will have a single vote! We demonstrate the robustness of \Name{} to poisoning through analytical proofs and  experimentation.  We also show \Name{}'s  high communication efficiency.
Our experiments demonstrate the superiority of \Name{} in real-world FL settings. In particular, \textbf{(1)} \Name{} is substantially more robust to poisoning attacks than state-of-the-art robust aggregation algorithms; \textbf{(2)} \Name{} achieves performances similar to the state-of-the-art federated averaging (FedAvg) with significantly lower communication costs, e.g., for CIFAR10, \Name{} achieves same performance as FedAvg with $\sim35\%$ lower communication cost.

\end{abstract}

\section{Introduction}
Federated Learning (FL) is an emerging learning paradigm, where mutually untrusted  \emph{clients} (e.g., Android devices) collaborate to train a shared model, called the \emph{global model}, without explicitly sharing their local training data. FL training involves a \emph{server} (e.g., a Google server) who repeatedly collects model updates that the clients compute using their local private data, aggregates the clients’ updates using an \emph{aggregation rule} (AGR), and finally uses the aggregated  updates to tune the jointly
trained model (called the \emph{global model}), which is broadcast to a subset of the clients at the end of each FL training round.
A major obstacle to the real-world adoption of FL in critical tasks is the threat of poisoning 
~\cite{kairouz2019advances,li2020federated,shejwalkar2021back}, which is the focus of our work.

\paragraphb{\em Robustness to poisoning attacks:} Most of the distributed learning algorithms, including FedAvg~\cite{mcmahan2017communication} and FedProx~\cite{li2018federated}, operate on mutually untrusted clients and server. This makes distributed learning susceptible to the threat of \emph{poisoning}~\cite{kairouz2019advances, shejwalkar2021back, bonawitztowards}. A \emph{poisoning adversary} can either own or control a few of FL clients, called \emph{malicious clients}, and instruct them to share malicious updates with the central server in order to reduce the performance of the global model. There are three approaches to poisoning FL: \emph{targeted}~\cite{bhagoji2019analyzing, sun2019can} attacks aim to reduce the utility of the global FL model on specific test inputs of adversary’s choice; \emph{untargeted}~\cite{baruch2019a, fang2020local,shejwalkar2021manipulating} attacks aim to reduce the utility of global model on arbitrary test inputs; and \emph{backdoor}~\cite{bagdasaryan2020backdoor,wang2020attack, xie2019dba} attacks aim to reduce the utility on test inputs that contain a specific signal called the trigger. 
In our work, we focus on the more severe threat of untargeted poisoning~\cite{shejwalkar2021back}, which, unlike targeted and backdoor poisoning, affects the majority FL clients.

\begin{figure}
    \begin{center}
    \includegraphics[width=0.5\textwidth]{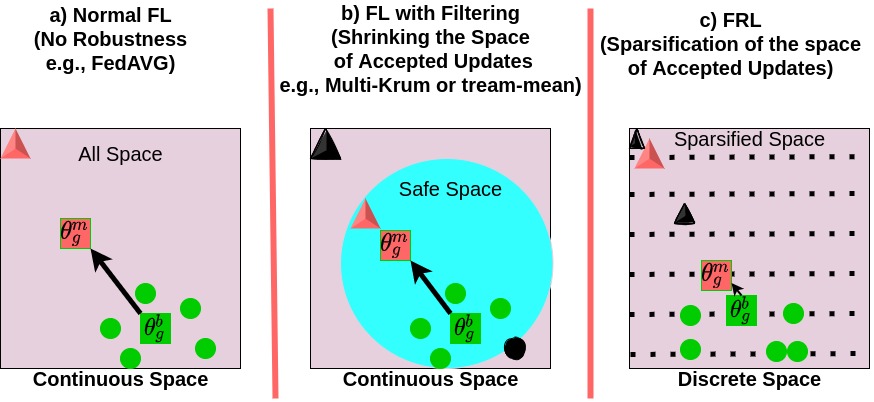}
    \end{center}
    \vspace{-15pt}
    \caption{The space of client updates. Green circles represent benign updates and red triangles represent malicious updates.
    To defend against poisoning updates, existing robust AGRs filter the updates by creating a safe space (continuous $\in$~$\mathbb{R}^d$). On the other hand, FRL limits the choices of clients by having a discrete space of updates (a permutation of integers $\in[1,d]$).
    $\theta_g^b$ (shown by green square) demonstrates the aggregated model for benign users, and $\theta_g^m$ (shown by the red square) demonstrates the aggregated model considering malicious updates. Black objects are updates that are ruled out by the server.} 
    \label{fig:shrinkAGR}
    \vspace{10pt}
\end{figure}

\paragraphb{\em High-level intuition of FL untargeted poisoning attacks: }
Figure~\ref{fig:shrinkAGR} shows how the poisoning adversary finds malicious updates in the space of possible updates which maximize the distance between benign and malicious aggregates.
When the server's aggregation rule (AGR) is not robust, e.g., dimension-wise average AGR~\cite{mcmahan2017communication}, there is no limitation on the adversary's choices, so they can maximize their goal using a malicious update that is arbitrarily far from benign updates; Figure~\ref{fig:shrinkAGR}-a) depicts this.
Therefore, even a single malicious client can jeopardize the accuracy of the global model trained using FedAvg~\cite{blanchard2017machine}. Current robust AGRs, such as Multi-krum~\cite{blanchard2017machine} or Trimmed-mean~\cite{YinCRB18} limit the space of acceptable updates, i.e., the safe zone shown in Figure~\ref{fig:shrinkAGR}-b). These robust AGRs only consider the updates that are in the safe zone and thereby reduce the adversary's choices of impactful malicious updates. 

\paragraphb{\em Continuous versus discrete space of updates:}
Figure~\ref{fig:shrinkAGR}-c) shows how our proposed defense (FRL, which is introduced next) limits the poisoning adversary's choices of malicious updates by making the space of acceptable updates \emph{discrete}. 
To the best of our knowledge, most of previous Byzantine robust FL algorithms use a \emph{continuous} space of updates ($\in$~$\mathbb{R}^d$), as their frameworks are built on exchanging trained (32-bit) weight parameters.
On the other hand, in our approach, the clients send their updates in the form of \emph{edge rankings}, i.e., a permutation of integers $\in[1,d]$ where $d$ is the size of the network layer; more useful edges have higher ranks.
In Figure~\ref{fig:shrinkAGR}-c), the black dots show the discrete space of acceptable client updates. For example, a network with 4 edges can have $4!$ possible permutations of edge rankings starting from [1,2,3,4] to [4,3,2,1]. On the other hand, in FL algorithms with a continuous space of updates (with or without a safe zone), the adversary's choices are 4 weight parameters (each of 32 bits). Note that, sparsification of the space of acceptable updates is different from sparsification of model updates used in compression methods, e.g., TopK~\cite{AlistarhH0KKR18}), RandomK~\cite{stich2018sparsified} and Sketched-SGD~\cite{ivkin2019communication}. In these methods, the FL client sends only a fraction of model updates instead of all of them, but each parameter still has a continuous space.

\paragraphb{\em \underline{F}ederated \underline{R}ank \underline{L}earning (\Name{}):}
We present \Name{}, a novel FL algorithm that concurrently achieves the two goals of robustness against poisoning attacks and communication efficiency. \Name{} uses a novel learning paradigm called \emph{supermasks} training~\cite{DBLP:conf/nips/ZhouLLY19, ramanujan2020what} to create edge rankings, which, as we will show, allows FRL to reduce communication costs while achieving significantly stronger robustness.
Specifically, in \Name{}, clients collaborate  to find a \emph{subnetwork} within a \emph{randomly initialized} neural network which we call the \emph{supernetwork} (this is in contrast to conventional FL where clients collaborate to \emph{train} a neural network).
The goal of training in \Name{} is to collaboratively rank the supernetwork's edges based on the importance of each edge and find a \emph{global ranking}. The global ranking can be converted to a supermask, which is a binary mask of 1's and 0's, that is superimposed on the random neural network (the supernetwork) to obtain the final  subnetwork. For example, in our experiments, the final subnetwork is constructed using the top 50\% of all edges.
The subnetwork is then used for downstream tasks, e.g., image classification, hence it is equivalent to the global model in conventional FL. Note that in entire \Name{} training, weights of the supernetwork \emph{do not} change.

%

More specifically, 
each  \Name{} client computes the importance of the edges of the supernetwork based on their local data. The importance of the edges is represented as a ranking vector. 
Each \Name{} client will use the \emph{edge popup} algorithm~\cite{ramanujan2020what} and their data to compute their local rankings (the edge popup algorithm aims at learning which edges in a supernetwork are more important over the other edges by minimizing the loss of the subnetwork on their local data). 
Each client then will send their local edge ranking to the server.
Finally, the \Name{} server uses a novel \emph{voting} mechanism to aggregate client rankings into a global ranking vector, which represents which edges of the random neural network (the supernetwork) will form the global subnetwork.

\paragraphb{\em Intuitions on \Name{}'s robustness:} 
In traditional FL algorithms, clients send  large-dimension model updates  $\in$~$\mathbb{R}^d$ (real numbers) to the server, providing  malicious clients significant flexibility in fabricating malicious updates.  By contrast, \Name{} clients merely share the rankings of the edges of the supernetwork, i.e.,  integers $\in[1,d]$, where $d$ is the size of the supernetwork. 
This allows the \Name{} server to use a voting mechanism to aggregate client updates (i.e., ranks), therefore, providing high resistance to adversarial ranks submitted by poisoning clients, since each client can only cast a single vote!  
Therefore, as we will show both theoretically and empirically, \Name{} provides robustness by design and reduces the impact of untargeted poisoning attacks. Furthermore, unlike most existing robust FL frameworks, \Name{} does not require any knowledge about the percentages of malicious clients.

\paragraphb{\em Intuitions on \Name{}'s communication efficiency:} 
In \Name{}, the clients and the server communicate just the rankings of the edges in the supernetwork, i.e., a  permutation of indices in $[1,d]$. Ranking vectors are generally significantly smaller than the global model. This, as we will show, significantly reduces the upload and download communication in \Name{} compared to Federated Averaging (FedAvg)~\cite{mcmahan2017communication}, where clients communicate model parameters, each of 32/64 bits. 


\paragraphb{\em Evaluation results:}
We experiment with three datasets in real-world heterogeneous FL settings and show that: \textbf{(1)} \Name{} achieves similar performance (e.g., model accuracy) as state-of-the-art FedAvg but with significantly reduced communication costs: for CIFAR10, the accuracy and communication cost per client are  85.4\% and 40.2MB for FedAvg, while 85.3\% and 26.2MB for \Name{}. \textbf{(2)} \Name{} is highly robust to poisoning attacks as compared to state-of-the-art robust aggregation algorithms: from 85.4\% in the benign setting,  10\% malicious clients reduce the accuracy of FL to 56.3\% and 58.8\% with Trimmed-Mean~\cite{YinCRB18} and Multi-Krum~\cite{blanchard2017machine}, respectively, while \Name{}'s performance only decreases to 79.0\%.

We also compare  \Name{} with two communication reduction methods, SignSGD~\cite{bernstein2018signsgd} and TopK~\cite{AlistarhH0KKR18} and show that
\Name{} produces comparable communication costs and model accuracies. For instance, 
on CIFAR10, \Name{}, SignSGD, and TopK achieve 85.3\%, 79.1\%, and 82.1\% test accuracy, respectively, when the corresponding communication costs (download and upload) are 26.2MB, 20.73MB, and 30.79MB. 
On the other hand, \Name{} offers significantly superior robustness. For instance, on CIFAR10, 10\% (20\%) malicious clients reduce the accuracy of SignSGD to 39.7\% (10.0\%), but \Name{}'s accuracy decreases to only 79.0\% (69.5\%). TopK is incompatible with existing robust aggregation algorithms, hence uses Average aggregation and is as vulnerable as FedAvg, especially in the real-world heterogeneous settings. 

In summary, we propose a federated learning approach that is built on exchanging rankings instead of parameter weights, and we show a ranking-based FL is more robust to untargeted poisoning attacks. Our key contributions are as follows:

\begin{compactitem}
\item We show that FL's robustness to poisoning can improve by sparsifying the space of updates sent by FL clients, therefore reducing the attacker's search space and enabling a voting-based aggregation. We particularly design  \textbf{F}ederated \textbf{R}ank \textbf{L}earning (\Name{}), a novel FL system in which clients collaboratively train a global model by ranking the importance of the edges of a random network based on their local data. 
    \item We evaluate \Name{} on three benchmark datasets including MNIST, CIFAR10, FEMNIST where we split them in heterogeneous fashion among a large number of users, i.e., among 1000, 1000, 3400 users respectively. We show that \Name{} provide more robustness and competitive communication efficiency compared to state-of-the-art Byzantine robust aggregation rules and compression techniques. 
    We evaluate the performance of \Name{} with two different methods of heterogeneous data distributions to consider most of the real-world non-iid data distributions.
    \item We derive theoretical robustness bounds of \Name{} that shows that \Name{} provides robustness by  design without any knowledge of the number of malicious clients. 

\end{compactitem}

\section{Related Works}
\paragraphb{Supermask Learning:}
Modern neural networks have a very large number of parameters. These networks are generally overparameterized~\cite{dauphin2013big, denil2013predicting, LotteryFL, li2021fedmask}, i.e., they have more parameters than they need to perform a particular task, e.g., classification.
The \emph{lottery ticket hypothesis}~\cite{frankle2018lottery} states that a  fully-trained neural network, i.e., \emph{supernetwork}, contains sparse \emph{subnetworks}, i.e., subsets of all neurons in supernetwork, which can be trained from scratch (i.e., by training same initialized weights of the subnetwork) and achieve performances close to the fully trained supernetwork.
The lottery ticket hypothesis allows for massive reductions in the sizes of neural networks. Ramanujan et al.~\cite{ramanujan2020what} offer a complementary conjecture that an overparameterized neural network with randomly initialized weights contains subnetworks which perform as good as the fully trained network.



\paragraphb{Poisoning Attacks and Defenses for Federated Learning (FL):}
FL involves mutually untrusting  clients. 
Hence, a \emph{poisoning adversary} may own or compromise some of the FL clients, called \emph{malicious clients}, with the goal of mounting a \emph{targeted} or \emph{untargeted} poisoning attack. In a targeted attack~\cite{bhagoji2019analyzing, sun2019can}, the goal is to reduce the utility of the model on specific test inputs, while in the untargeted attack~\cite{baruch2019a, fang2020local,shejwalkar2021manipulating}, the goal is to reduce the utility for all (or most)  test inputs. It is shown~\cite{blanchard2017machine} that even a single malicious client can mount an effective untargeted attack on FedAvg. 

In order to make FL robust to  the presence of such malicious clients, the literature has designed various \emph{robust aggregation rules (AGR)}~\cite{blanchard2017machine, DBLP:conf/icml/MhamdiGR18, YinCRB18,chang2019cronus}, which aim to remove or attenuate the updates that are more likely to be malicious according to some criterion. For instance, Multi-krum~\cite{blanchard2017machine} repeatedly removes updates that are far from the geometric median of all the updates, and Trimmed-mean~\cite{YinCRB18} removes the largest and smallest values of each update dimension and calculates the mean of the remaining values.
%
Unfortunately, these robust AGRs are not very effective in  non-convex FL settings and multiple works have demonstrated strong targeted~\cite{wang2020attack,bhagoji2019analyzing} and untargeted attacks~\cite{shejwalkar2021manipulating,fang2020local} on them.



\paragraphb{Communication Cost of FL:}
In many real-world applications of FL, it is essential to minimize the communication between FL server and clients. Especially in cross-device FL, the clients (e.g., mobile phones and wearable devices) have limited resources and communication can be a major bottleneck. 
%
There are two major types of communication reduction methods: (1) \emph{Qunatization} methods reduce the resolution of (i.e., number of bits used to represent) each dimension of a client update. For instance, SignSGD~\cite{bernstein2018signsgd} uses the sign (1 bit) of each dimension of model updates. 
(2) \emph{Sparsification} methods propose to use only a subset of all the update dimensions. For instance, in TopK~\cite{AlistarhH0KKR18}, only the largest K\% update dimensions are sent to the server in each FL round.
We note that, communication reduction methods primarily focus on and succeed at reducing upload communication (client $\rightarrow$ server), but they use the entire model in download communication (server $\rightarrow$ client).

\section{Preliminaries}

\subsection{Federated Learning}
In FL~\cite{mcmahan2017communication, kairouz2019advances, konevcny2016federated}, 
$N$ clients collaborate to train a global model  without directly sharing their data. 
In  round $t$, the service provider (server) selects $n$ out of $N$ total clients and  sends them the most recent global model $\theta^t$. Each client trains a local model for $E$ local epochs on their data starting from the $\theta^t$ using stochastic gradient descent (SGD). Then the client  sends back the calculated gradients ($\triangledown_k$ for $k$th client) to the server. The server then aggregates the collected gradients and  updates the global model for the next round.
FL can be either cross-device or cross-silo~\cite{kairouz2019advances}. In cross-device FL, N is large (from few thousands to billions) and only a
small fraction of clients is chosen in each FL training round, i.e., $n \ll N$. By contrast, in cross-silo FL, N is moderate (up to 100) and all clients are chosen in each round, i.e., $n = N$. 
In this work, we evaluate the performance of \Name{} and other FL baselines for cross-device FL under realistic production FL settings.


\subsection{Edge-popup Algorithm}
The edge-popup (EP) algorithm~\cite{ramanujan2020what} is a novel optimization method to find supermasks within a large, randomly initialized neural network, i.e., a supernetwork, with performances close to the fully trained supernetwork.
 EP algorithm does not train the weights of the network, instead only decides  the set of edges to keep and removes the rest of the edges (i.e., pop).
Specifically, EP algorithm assigns a positive score to each of the edges in the supernetwork. On forward pass, it selects top k\% edges with highest scores, where k is the percentage of the total number of edges in the supernetwork that will remain in the final subnetwork. On the backward pass, it updates the scores with the straight-through gradient estimator~\cite{bengio2013estimating}.

Algorithm~\ref{alg:edgepop} presents EP algorithm.
Suppose in a fully connected neural network, there are $L$ layers and layer $\ell\in[1,L]$ has $n_{\ell}$ neurons, denoted by $V^{\ell}=\{V_1^{\ell}, ..., V_{n_{\ell}}^{\ell}\}$. If $I_{v}$ and $Z_v$ denote the input and output for neuron $v$ respectively, then the input of the node $v$ is the weighted sum of all nodes in previous layer, i.e., $I_v=\sum_{u \in V^{\ell-1}} W_{uv}Z_{u}$.
Here, $W_{uv}$ is the weight of the edge connecting $u$ to $v$. Edge-popup algorithm tries to find subnetwork $E$, so the input for neuron $v$ would be: $I_v=\sum_{(u,v) \in E} W_{uv}Z_{u}$.

\paragraphb{Updating scores.} Consider an edge $E_{uv}$ that connects two neurons $u$ and $v$,  $W_{uv}$ be the weight of $E_{uv}$, and $s_{uv}$ be the score assigned to the edge $E_{uv}$ by Edge-popup algorithm. 
Then the edge-popup algorithm removes edge $E_{uv}$ from the supermask if its score $s_{uv}$ is not high enough. Each iteration of supermask training updates the scores of all edges such that, if having an edge $E_{uv}$ in subnetwork reduces loss (e.g., cross-entropy loss) over training data, the score $s_{uv}$ increases.

        \begin{algorithm}[H] 
            \caption{Edge-popup (EP) algorithm: it finds a subnetwork of size $k$\% of the entire network $\theta$} \label{alg:edgepop}
            \begin{algorithmic}[1] 
            \State \textbf{Input:} number of local epochs $E$, training data $D$, initial weights $\theta^w$ and scores $\theta^s$, subnetwork size $k$\%, learning rate $\eta$
            \For{$e \in [E]$}
            \State $\mathcal{B}\gets$ Split $D$ in $B$ batches
            \For{batch $b\in[B]$}
                \State \textsc{EP Forward} ($\theta^w, \theta^s, k, b$)
                \State $\theta^s=\theta^s - \eta\nabla \ell(\theta^s;b)$
            \EndFor
        \EndFor
        \State \textbf{return} $\theta^s$
                \Function{EP forward}{$\theta^w, \theta^s, k, b$}
                \State $m \gets$ sort$(\theta^s)$
                \State $t \gets$ $int((1-k)* len(m))$
                \State $m[:t]=0$
                \State $m[t:]=1$
                \State $\theta^p =\theta^w \odot \mathbf{m}$
                \State \textbf{return} $\theta^p(b)$
                \EndFunction
                
            \end{algorithmic}
        \end{algorithm}

The algorithm selects top k\% edges (i.e., finds a subnetwork with sparsity of k\%) with highest scores, so $I_v$ reduces to $I_v=\sum_{u \in V^{\ell-1}} W_{uv} Z_{u} h(s_{uv})$ 
where $h(.)$ returns $1$ if the edge exists in top-k\% highest score edges and $0$ otherwise. Because of existence of $h(.)$, which is not differentiable, it is impossible to compute the gradient of loss with respect to $s_{uv}$.
Recall that, the Edge-popup algorithm use straight-through gradient estimator~\cite{bengio2013estimating} to compute gradients. In this approach, $h(.)$ will be treated as the identity in the backward pass meaning that the upstream gradient (i.e., $\frac{\partial L}{\partial I_v}$) goes straight-through $h(.)$. Now using chain rule, we can derive $\frac{\partial L}{\partial I_v} \frac{\partial I_v}{\partial s_{uv}} = \frac{\partial L}{\partial I_v} W_{uv}Z_{u}$
%
 where $L$ is the loss to minimize. Then we can SGD with step size $\eta$ to update scores as 
 $s_{uv} \xleftarrow{} s_{uv} - \eta \frac{\partial L}{\partial I_v} Z_u W_{uv}$.

\section{Federated Rank Learning: Design}\label{fsl:design}

\begin{algorithm*}
\caption{Federated Ranking Learning (\Name{})}\label{alg:FSL+}
\begin{algorithmic}[1]
\State \textbf{Input:} number of FL rounds $T$, number of local epochs $E$, number of selected users in each round $n$, seed \textsc{seed}, learning rate $\eta$, subnetwork size $k$\%

\State \serverbox{Server: Initialization}
\State $\theta^s, \theta^w \gets$ Initialize random scores and weights of global model $\theta$ using \textsc{seed}
\State $R_{g}^{1} \gets \textsc{ArgSort}(\theta^s)$ \Comment{Sort the initial scores and obtain initial rankings}

\For{$t \in [1,T]$}
    \State $U \gets$ set of $n$ randomly selected clients out of $N$ total clients
    \For{$u$ in $U$}
        \State \clientbox{Clients: Calculating the ranks}
        \State $\theta^s, \theta^w \gets $ Initialize scores and weights using \textsc{seed}
        \State $\theta^s[R_{g}^{t}] \gets \textsc{sort}(\theta^s)$ \Comment{sort the scores based on the global ranking}
        \State $S \gets$ Edge-PopUp($E, D_u^{tr}, \theta^w, \theta^s, k, \eta$) \Comment{Client u uses Algorithm\ref{alg:edgepop} to train a supermask on its local training data}
        \State $R_{u}^{t} \gets \textsc{ArgSort}(S)$ \Comment{Ranking of the client}
    \EndFor
    \State \serverbox{Server: Majority Vote}
    \State $R_{g}^{t+1} \gets \textsc{Vote}(R_{\{u \in U\}}^{t})$ \Comment{Majority vote aggregation}
\EndFor
\Function{Vote}{$R_{\{u \in U\}}$ }:
    \State $V \gets \textsc{ArgSort}(R_{\{u \in U\}})$
    \State $A \gets \textsc{Sum}(V)$
    \State \textbf{return} $\textsc{ArgSort}(A)$
\EndFunction
\end{algorithmic}
\end{algorithm*}

\begin{figure*}[h]
\begin{minipage}[t]{0.65\textwidth}
\includegraphics[width=1.0\textwidth]{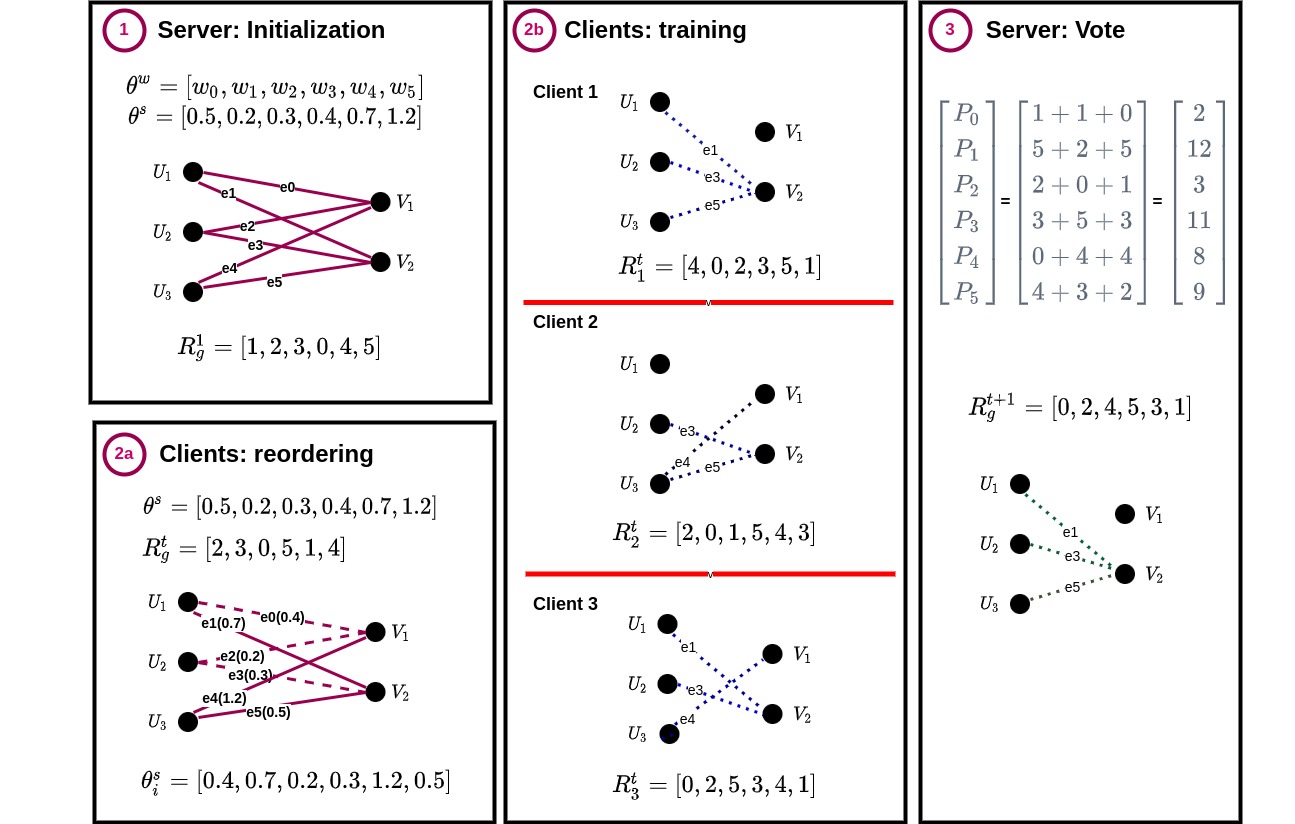}
\vspace*{-10pt}
\caption{A single \Name{} round with three clients and network of 6 edges. } \label{fig:fsl_flow}
\end{minipage}
\begin{minipage}[t]{0.35\textwidth}
\includegraphics[width=1.0\textwidth]{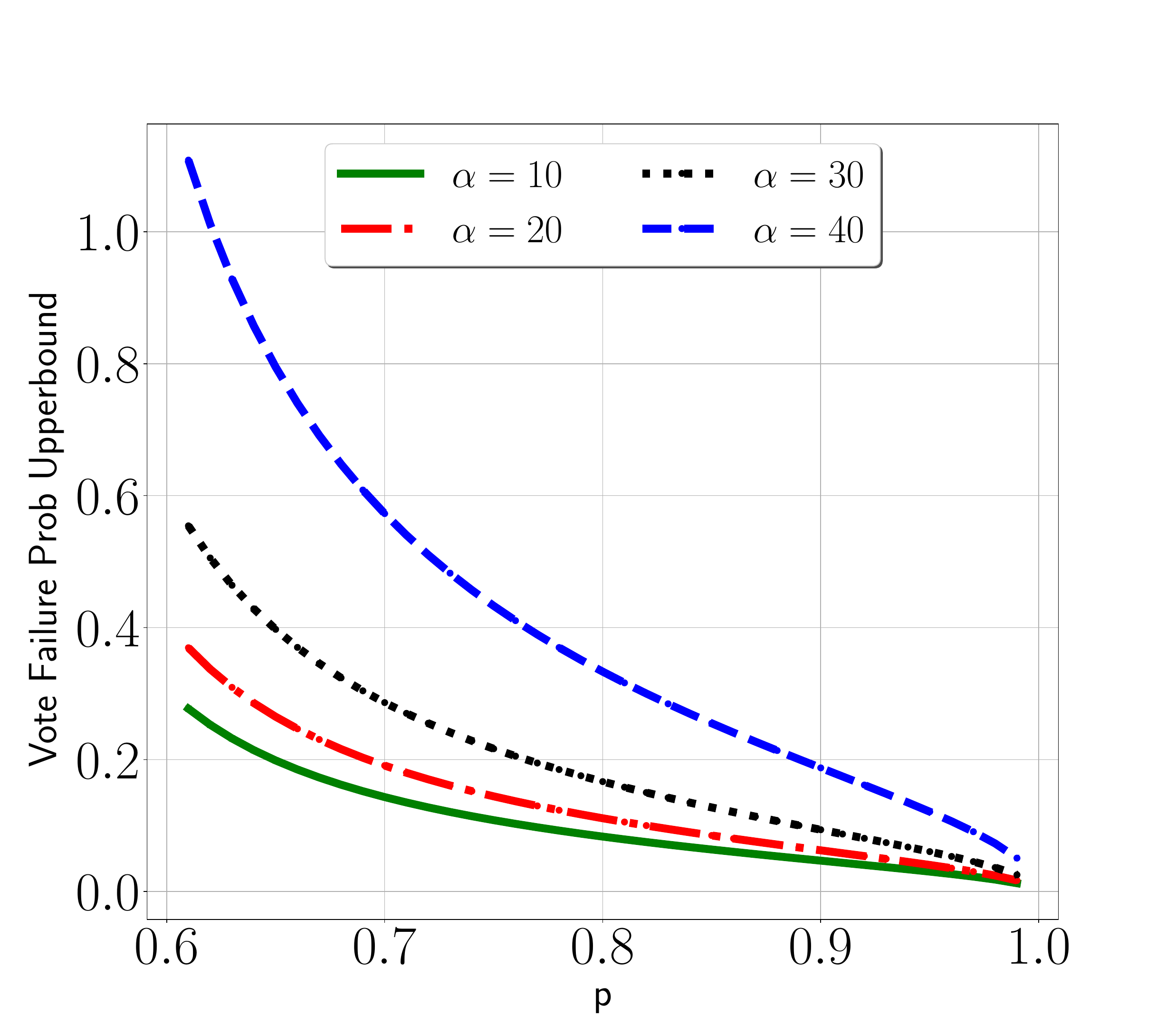}
\vspace*{-10pt}
\caption{Upper bound on the failure probability of \textsc{Vote(.)} function in \Name{}.} \label{fig:upper}
\end{minipage}
\end{figure*}

In this section, we provide the design of our federated rank learning (\Name{}) algorithm. 
\Name{} clients collaborate (without sharing their local data) to \emph{find a subnetwork} within a randomly initialized, untrained neural network called the \emph{supernetwork}.
Algorithm~\ref{alg:FSL+} describes  \Name{}'s training. Training a global model in \Name{} means to first find a  unanimous ranking of supernetwork edges and then use the subnetwork of the top ranked edges as the final output.

\paragraphb{\Name{} objective:} The optimization problem of \Name{} is to find a global ranking $R_{g}$ which produces a global binary mask $m$ that minimizes the average loss of all of clients with that subnetwork $(\theta^w \odot \mathbf{m})$. \Name{} aims to solve:

\begin{align}
    \min_{R_{g}} F(\theta^w, R_{g}) = \min_{R_{g}} \sum_{i=1}^N \lambda_i L_i(\theta^w \odot \mathbf{m}) \\ \nonumber 
    \; \; \text{s.t.}  \; \;\mathbf{m}[R_{g}<k]=0 \; \; \text{and} \; \; \mathbf{m}[R_{g}\geq k]=1 
\end{align}

where $N$ is the total number of participating clients and $L_i$ is the loss function for the $i$th client. $\lambda_i$ shows the importance of the $i^{th}$ client in empirical risk minimization; $\lambda_i=\frac{1}{N}$ gives same importance to all the participating clients. $\mathbf{m}$ is the final mask that contains the edges of top $k$ ranks, i.e., edges in top $k$ ranks (layer-wise) get '1' in the binary mask, and others get '0' in the mask. $\theta^w \odot \mathbf{m}$ shows the subnetwork inside the random and fixed weights $\theta^w$ that all clients unanimously vote for. 
In Appendix~\ref{app:optimization}, we show how \Name{} minimizes its objective.  

We detail a round of \Name{} training and depict it in Figure~\ref{fig:fsl_flow}, where we use a supernetwork with six edges $e_{i\in[0,5]}$ to demonstrate a single \Name{} round and consider three clients $C_{j\in[1,3]}$ who aim to find a subnetwork of size $k$=50\% of the original supernetwork.






\subsection{Server: Initialization Phase (Only for round $t=1$)}

In the first round, the \Name{} server chooses a random seed \textsc{Seed} to generate initial random weights $\theta^w$  and scores $\theta^s$ for the global supernetwork $\theta$; note that, $\theta^w$, $\theta^s$, and \textsc{Seed} remain constant during the entire \Name{} training. Next, the \Name{} server shares  \textsc{Seed} with  \Name{} clients, who can then locally reconstruct the initial weights $\theta^w$ and scores $\theta^s$ using \textsc{Seed}. 
Figure~\ref{fig:fsl_flow}-\circled{1} depicts this step. 

Recall that, the goal of \Name{} training is to find the most important edges in $\theta^w$ without changing the weights.
Unless specified otherwise, both server and clients use the Signed Kaiming Constant algorithm~\cite{ramanujan2020what} to generate random weights and the Kaiming Uniform algorithm~\cite{he2015delving} to generate random scores.
However, in Section~\ref{Exp:init}, we also explore the impacts of  different weight initialization algorithms
on the performance of \Name{}.  We use the same seed to initialize weights and scores.

At the beginning, the \Name{} server finds the global rankings of the initial random scores (Algorithm~\ref{alg:FSL+} line 4), i.e., $R_g^1= \textsc{ArgSort($\theta^s)$}$.
We define \emph{rankings of a vector} as the indices of elements of vector when the vector is sorted from low to high, which is computed using \textsc{ArgSort} function.
%

\subsection{Clients: Calculating the ranks (For each round $t$)}

In the $t^{th}$ round, \Name{} server randomly selects $n$ clients among total $N$ clients, and shares the global rankings $R_{g}^{t}$ with them. 
Each of the selected $n$ clients locally reconstructs the weights $\theta^w$'s and scores $\theta^s$'s using \textsc{seed} (Algorithm~\ref{alg:FSL+} line 9). 
Then, each \Name{} client reorders the random scores based on the global rankings, $R_{g}^{t}$ (Algorithm~\ref{alg:FSL+} line 10); we depict this in Figure~\ref{fig:fsl_flow}-\circled{2a}.

Next, each of the $n$ clients uses reordered $\theta^s$ and finds a subnetwork within $\theta^w$ using Algorithm~\ref{alg:edgepop}; to find a subnetwork, they use their local data and $E$ local epochs (Algorithm~\ref{alg:FSL+} line 11). Note that, each iteration of Algorithm~\ref{alg:edgepop} updates the scores $\theta^s$.
Then client $u$ computes their local rankings $R_{u}^t$ using the final updated scores ($S$) and \textsc{Argsort}(.), and sends $R_{u}^t$ to the server.
%
Figure~\ref{fig:fsl_flow}-\circled{2a} shows how each of the selected $n$ clients reorders the random scores using global rankings. For instance, the initial global rankings for this round are $R^t_g=[2,3,0,5,1,4]$, meaning that edge $e_4$ should get the highest score ($s_4=1.2$), and edge $e_2$ should get the lowest score ($s_2=0.2$).

Figure~\ref{fig:fsl_flow}-\circled{2b} shows, for each client, the scores and rankings they obtained after finding their local subnetwork.
For example, rankings of client $C_1$  are $R^t_1=[4,0,2,3,5,1]$, i.e., $e_4$ is the least important and $e_1$ is the most important edge for $C_1$. Considering desired subnetwork size to be 50\%, $C_1$ uses edges \{3,5,1\} in their final subnetwork.

\subsection{Server: Majority Vote  (For each round $t$)}\label{fsl:design:vote}
The server receives all the local rankings of the selected $n$ clients, i.e., $R^t_{\{u \in U\}}$. 
Then, it performs a majority vote over all the local rankings using \textsc{Vote(.)} function. 
Note that, for client $u$, the index $i$ represents the  importance of the edge ${R^t_u[i]}$ for $C_u$.
For instance, in Figure~\ref{fig:fsl_flow}-\circled{2b}, rankings of $C_1$ are $R^t_1=[4,0,2,3,5,1]$, hence the edge $e_4$ at index=0 is the least important edge for $C_1$, while the edge $e_1$ at index=5 is the most important edge.
Consequently, \textsc{Vote(.)} function assigns reputation=0 to edge $e_4$, reputation=1 to $e_0$, reputation=2 to $e_2$, and so on. In other words, for rankings  $R^t_u$ of $C_u$ and edge $e_i$,  \textsc{Vote(.)} computes the reputation of $e_i$ as its index in $R^t_u$. 
Finally, \textsc{Vote(.)} computes the total reputation of $e_i$ as the sum of reputations from each of the local rankings.
In Figure~\ref{fig:fsl_flow}-\circled{2b}, reputations of $e_0$ are 1 in $R^t_1$, 1 in $R^t_2$, and 0 in $R^t_3$, hence, the total reputation of $e_0$ is 2. We depict this in Figure~\ref{fig:fsl_flow}-\circled{3}; here, the final total reputations for edges $e_{\{i\in[0,5]\}}$ are $A=[2,12,3,11,8,9]$.
Finally, the server computes global rankings $R^{t+1}_g$ to use for round $t+1$ by sorting the final total reputations of all edges, i.e., $R^{t+1}_g=\textsc{Argsort}(A)$.

Note that,  \emph{all  \Name{} operations that involve sorting, reordering, and voting are performed in a layer-wise manner}. For instance, the server computes global rankings $R^t_g$ in round $t$ for each layer separately, and consequently, the clients selected in round $t$  reorder their local randomly generated scores $\theta^s$ for each layer separately.




\section{\Name{}'s Robustness to Poisoning}

\Name{} is a distributed learning algorithm with mutually untrusting clients. 
Hence, a \emph{poisoning adversary} may own or compromise some of \Name{} clients, called \emph{malicious clients}, and mount a \emph{targeted} or \emph{untargeted} poisoning attack. 
%
In our work, we consider the untargeted attacks as they are more severe than targeted attacks and can cause denial-of-service for all collaborating clients~\cite{shejwalkar2021back}, and show that \Name{} is secure against such poisoning attacks by design.

\paragraphb{Intuition on \Name{}'s robustness:}
Existing FL algorithms, including  robust FL algorithms, are shown to be  vulnerable to targeted and untargeted poisoning attacks~\cite{shejwalkar2021back} where  malicious clients corrupt the global model by sharing malicious model updates. 

One of the key reasons behind the susceptibility of existing algorithms  is that their model updates can have a large continuous space of values. For instance, to manipulate vanilla FedAvg, malicious clients send very large updates~\cite{blanchard2017machine}, and to manipulate Multi-krum and Trimmed-mean, \cite{fang2020local,shejwalkar2021manipulating} propose to perturb a benign update in a specific malicious direction.
On the other hand, in \Name{},  clients must send a permutation of indices $\in [1, n_{\ell}]$ for each layer. Hence, \Name{} significantly reduces the space of the possible malicious updates that an adversary can craft. Majority voting in \Name{} further reduces the chances of successful attack. Intuitively, this makes \Name{} design robust to poisoning attacks. Below, we make this intuition more concrete.


\paragraphb{The worst-case untargeted poisoning attack on \Name{}:}
Here, the poisoning adversary aims to reduce the accuracy of the final global \Name{} subnetwork on most test inputs.
To achieve this, the adversary should replace the high ranked edges   with low ranked edges in the final subnetwork.
%
For the worst-case analysis of \Name{}, we assume a very strong adversary (i.e., threat model): 1) each of the malicious clients has some data from benign distribution; 2) malicious clients know the entire \Name{} algorithm and its parameters; 3) malicious clients can collude.
Under this threat model we design a worst case attack on \Name{} (Algorithm~\ref{alg:poison}), which executes as follows: First, malicious clients compute rankings on their benign data and use \textsc{Vote}(.) algorithm to compute an aggregate rankings. Finally, each of the malicious clients uses the reverse of the aggregate rankings to share with the \Name{} server in given round. The adversary should invert the rankings layer-wise as the \Name{} server will aggregate the local rankings per layer too, and it is not possible to mount a model-wise attack.

\begin{algorithm}
\caption{\Name{} Poisoning}\label{alg:poison}
\begin{algorithmic}[1]
\State \textbf{Input:} 
number of malicious clients $M$, number of malicious local epochs $E'$, seed \textsc{seed}, global ranking $R_{g}^t$, learning rate $\eta$, subnetwork size $k$\%

\Function{Maliciousupdate}{$M, \textsc{seed}, R_g^t, E', \eta, k$}:
    \For{$mu \in [M]$} 
        \State \clientbox{Malicious Client Executes:}
        \State $\theta^s, \theta^w \gets $ Initialize scores and weights using \textsc{seed}
        \State $\theta^s[R_{g}^{t}] \gets \textsc{sort}(\theta^s)$

        \State $S \gets$ Edge-PopUp($E', D_u^{tr}, \theta^w, \theta^s, k, \eta$) 
        
        
        \State $R_{mu}^{t} \gets \textsc{ArgSort}(S)$  \Comment{Ranking of the malicious client}
    \EndFor
    \State \serverbox{Aggregation:}
    \State $R_{m}^{t} \gets \textsc{Vote}(R_{\{mu \in [M]\}}^{t})$ \Comment{Majority vote aggregation}
    \State \textbf{return} $\textsc{Reverse}(R_{m}^{t})$ \Comment{reverse the ranking}
\EndFunction
\end{algorithmic}
\end{algorithm}

Now we justify why the attack in Algorithm~\ref{alg:poison} is the worst case attack on \Name{} for the strong threat model we consider. Note that, \Name{} aggregation, i.e., \textsc{Vote}(.), computes the reputations using clients' rankings and sums the reputations of each network edge.
Therefore, the strongest poisoning attack would want to reduce the reputation of good edges. This can be achieved following the aforementioned procedure of Algorithm~\ref{alg:poison} to reverse the rankings computed using benign data.


\paragraphb{Theoretical analysis of robustness of \Name{} algorithm:}
In this section, we prove an upper bound on the failure probability of robustness of \Name{}, i.e., the probability that a good edge will be removed from the final subnetwork when malicious clients mount the worst case attack.

Following the work of~\cite{bernstein2018signsgd}, we make two assumptions in order to facilitate a concrete robustness analysis of \Name{}: a) each malicious client has access only to its own data,
and b) we consider a simpler \textsc{Vote}(.) function, where the \Name{} server puts an edge $e_i$ in the final subnetwork if more than half of the clients have $e_i$ (a good edge) in their local subnetworks. In other words, the rankings that each client sends to the server is just a bit mask showing that each edge should or should not be in the final subnetwork. The server makes a majority vote on the bit masks, and if an edge has more than half votes, it will be in the global subnetwork. {Our \textsc{Vote}(.) mechanism has more strict robustness criterion, as it uses more nuanced reputations of edges instead of bit masks.
Hence, the  upper bound on failure probability in this section also applies to the \Name{} \textsc{vote}(.) function.}
%

The probability that our voting system fails is the probability that more than half of the votes do not include  $e_i$ in their subnetworks. The upper bound on the probability of failure would be $1/2 \sqrt{\frac{np(1-p)}{(n(p +\alpha(1 -2 p) - 1/2))^2}}$, where  $n$ is the number of clients being processed,  $p$ shows the probability that a benign client puts $e_i$ in their top ranks, and  $\alpha$ is the fraction of malicious clients. Due to space limitations, we defer the detailed proof to  Appendix~\ref{app:Robust1}. 
Figure~\ref{fig:upper} shows the upper bound on the failure of \textsc{Vote(.)} for different values of $\alpha$ and $p$. We note that, the higher the probability $p$, the higher the robustness of \Name{}.

\section{\Name{}'s Communication Efficiency} \label{sec: SFSL}
In FL, and especially in the cross-device setting,  clients have limited communication bandwidth. Hence, FL algorithms  must be communication efficient. We discuss here the communication cost of \Name{} algorithm.
In the first round, the \Name{} server only sends one seed of 32 bits
to all the \Name{} clients, so they can construct the random weights ($\theta^w$) and scores ($\theta^s$). In a round $t$, each of selected \Name{} clients receives the global rankings $R^t_g$ and sends back their local rankings $R^t_u$. The rankings are a permutation of the indices of the edges in each layer, i.e., of $[0,n_{\ell}-1] \forall \ell \in [L]$ where $L$ is the number of layers and $n_{\ell}$ is the number of parameters in $\ell$th layer.

We use the naive approach to communicate layer-wise rankings, where each \Name{} client exchanges a total of $\sum_{\ell \in [L]} n_{\ell} \times \log(n_{\ell})$ bits. Because, for the layer $\ell$, the client receives and sends $n_{\ell}$ ranks where each one is encoded with $\log(n_{\ell})$ bits.  On the other hand, a client exchanges $\sum_{\ell \in [L]} n_{\ell}\times 32$ bits in FedAvg, when 32 bits are used to represent each of $n_{\ell}$ weights in layer $\ell$. 
In Section~\ref{Exp:comp}, we measure the performance and communication cost of \Name{} with other existing FL compressors SignSGD~\cite{bernstein2018signsgd} and TopK~\cite{DBLP:conf/emnlp/AjiH17,AlistarhH0KKR18}. 
%



\begin{figure}
    \begin{center}
    \includegraphics[width=0.5\textwidth]{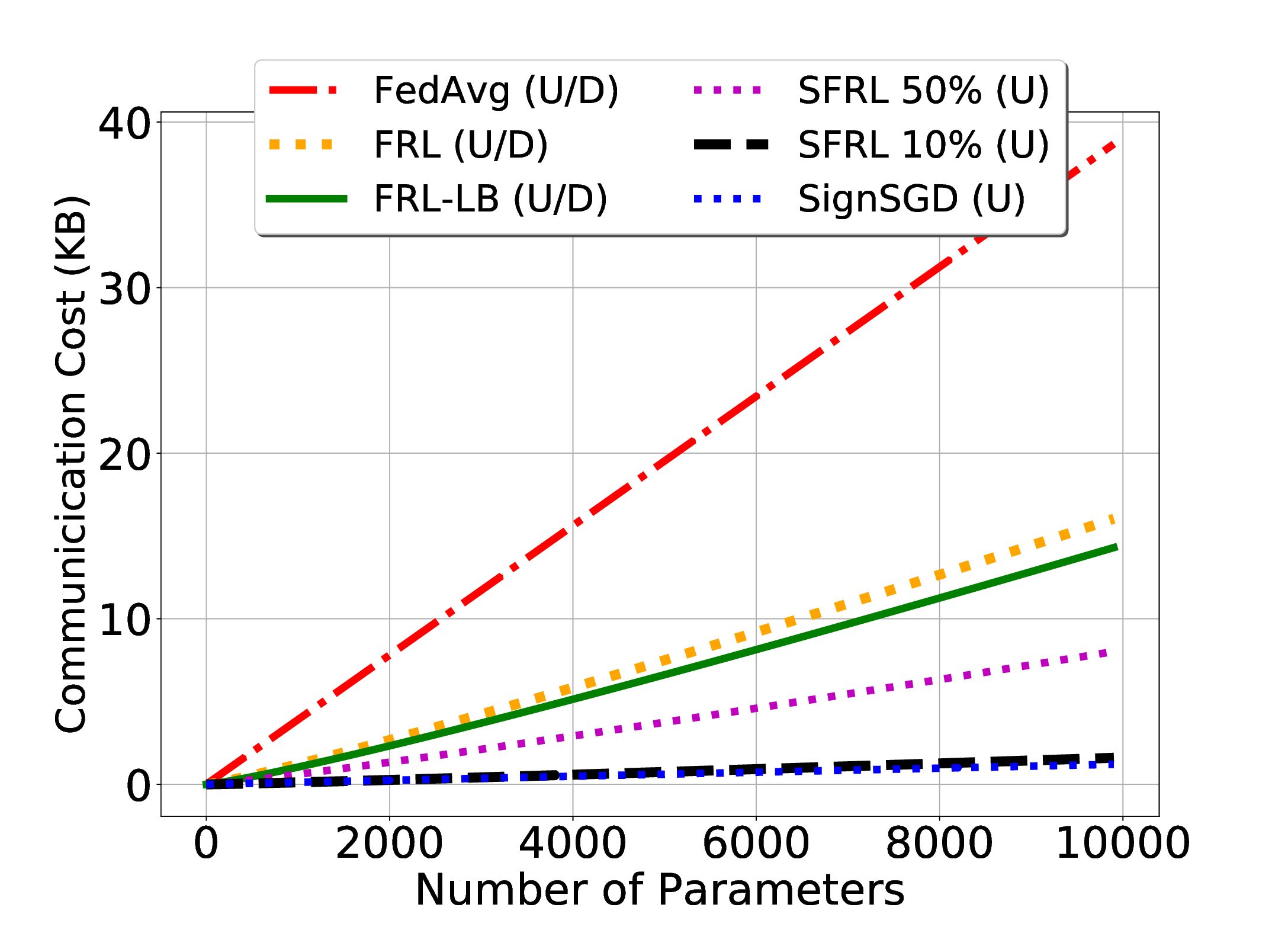}
    \end{center}
    \vspace{-15pt}
    \caption{ Communication cost Analysis. U and D are showing upload and download communication costs. Please note that the download communication cost of all S\Name{}s would be the same as \Name{}. Download communication cost of SignSGD would be the same as FedAvg too.}
    \label{fig:comm}
    \vspace{10pt}
\end{figure}




\paragraphb{Sparse-\Name{}:}
Here, we propose Sparse-\Name{}, a simple extension of \Name{} to further reduce the communication cost.
In Sparse-\Name{}, a client sends only the most important ranks of their local rankings to the server for aggregation. For instance, in Figure~\ref{fig:fsl_flow}, client $C_1$ sends $R^t_1=[4,0,2,3,5,1]$ in case of \Name{}. But in sparse-\Name{}, with sparsity set to 50\%, client $C_1$ sends just the top 3 rankings, i.e., sends $R'^t_1=[3,5,1]$.
For each client, the sparse-\Name{} server assumes 0 reputation for all of the edges not included in the client's rankings, i.e., in Figure~\ref{fig:fsl_flow}, sparse-\Name{} server will assign reputation=0 for edges $e_4$, $e_0$, and $e_2$.
Then the server uses \textsc{Vote}(.) to compute total reputations of all edges and then sort them to obtain the final aggregate global rankings, i.e., $R^{t+1}_g$, to use for subsequent rounds. 
We observe in out experiments, that sparse-\Name{} performs very close to \Name{}, even with sparsity as low as 10\%, while also significantly reducing the communication cost.


\paragraphb{Lower-bound of communication cost of \Name{}:}Since the \Name{} clients send and receive layer-wise rankings of indices, i.e., integers $\in[0, n_{\ell}-1]$, for layer $\ell$, 
there are $n_{\ell}!$ possible permutations for layer $\ell \in [L] $. If we use the best possible compression method in \Name{}, an \Name{} client needs to send and receive $\sum_{\ell \in [L]} \log(n_{\ell}!)$ bits. Therefore, the download and upload bandwidth for each \Name{} client would be $\sum_{\ell \in [L]}  \log\left(n_{\ell} * (n_{\ell}-1) * ... *2 *1\right)=\sum_{\ell \in [L]} \sum_{i=1}^{n_{\ell}} log(i)$ bits. Please note that in our experiment, \Name{} clients send and receive the rankings without any further compression, and $\sum_{\ell \in [L]} \sum_{i=1}^{n_{\ell}} log(i)$ just shows a lower-bound of communication cost of \Name{}.

In Figure~\ref{fig:comm}, we compare the upload and download communication costs of one client per FL round for FedAvg, SignSGD, and different variants of \Name{} for different number of parameters. U and D are showing upload and download communication cost. Please note that the download communication cost of all S\Name{}s would be the same as \Name{}, and download communication cost of SignSGD would be similar to FedAvg too.
If we use a compression method to compress the local rankings (for upload) and global rankings (for download), we can improve communication cost of \Name{} to its lower-bound (FRL-LB in Figure~\ref{fig:comm}).
In this Figure, we can see that S\Name{} can provide competitive upload communication cost and lower download communication cost compared to SignSGD when the clients are sending only 10\% of top local rankings (S\Name{} 10\%). 


\section{Experiments}\label{Exp}
In this section, we investigate the utility, robustness, and communication cost of our \Name{} framework. We use MNIST, CIFAR10, and FEMNIST data and  distribute them in non-iid fashion among 1000, 1000, and 3400 clients respectively.  
We run all the experiments for 2000 global rounds of \Name{} and FL and select 25 clients in each round.
At the end of the training, we calculate the test accuracy of all the clients on the final global model, and we report the mean and standard deviation of all clients' test accuracies in our experiments.

\paragraphb{Using Dirichlet distribution:}
Considering the heterogeneous data in the real-word cross-device FL, similar to previous works~\cite{reddi2020adaptive, hsu2019measuring}, we distribute MNIST and CIFAR10 among 1,000 clients in a non-iid fashion using Dirichlet distribution with parameter $\beta=1$. Note that increasing  $\beta$ results in more iid datasets.
Next, we split datasets of each client into training ($80\%$) and test ($20\%$).

In addition to \Name{}, we also evaluate Sparse-\Name{} in different settings. We use S\Name{} top x\% to denote a Sparse-\Name{} clients who sends top x\% of ranks in each round.

This section is organized as follows: in Section~\ref{sec:setupp} we discuss the experimental setup. In Section~\ref{Exp:security} we investigate the robustness and utility of \Name{}, followed by Section~\ref{Exp:comp} in which we discuss the communication cost of \Name{}. In Section~\ref{sec:noniid2} we show the effectiveness of \Name{} with respect to different methods of non-iid-ness. We conclude with Section~\ref{Exp:init} and Section~\ref{sec:size} where we study the effect of weight initialization and size of subnetwork in \Name{}.



\begin{table*}[ht]
\caption{Comparing the robustness of various FL algorithms: \Name{} and Sparse-\Name{} (S\Name{}) (in \textbf{bold}) outperform the state-of-the-art robust AGRs and SignSGD against the strongest of untargeted poisoning attacks.} \label{tab:mal}
\centering
\begin{tabular} {|c|c||c|c|c|}
  \hline
  Dataset & AGR & No malicious & 10\% malicious & 20\% malicious \\ \hline \hline
  \multirow{6}{*}{{\vtop{\hbox{\strut MNIST + LeNet}\hbox{\strut 1000 clients}}}} & FedAvg  & 98.8 (3.2) & 10.0 (10.0) & 10.0 (10.0) \\ \cline{2-5} 
  & Trimmed-mean  & 98.8 (3.2) &  95.1 (7.7) & 87.6 (9.5) \\ \cline{2-5} 
  & Multi-krum  & 98.8 (3.2) & 98.6 (3.3) & 97.9 (4.1)  \\ \cline{2-5} 
  & SignSGD  & 97.2 (4.6) & 96.6 (5.0) & 96.2 (5.6) \\ \cline{2-5} 
  & \textbf{\Name{}}  & \textbf{98.8 (3.1)} & \textbf{98.8 (3.1)} & \textbf{98.7 (3.3)} \\ \cline{2-5} 
  & \textbf{S\Name{} Top 50\%}  & \textbf{98.2 (3.8)} & \textbf{97.04 (4.4)} & \textbf{95.1 (7.8)} \\ 
\hline \hline

  \multirow{6}{*}{{\vtop{\hbox{\strut CIFAR10 + Conv8}\hbox{\strut 1000 clients}}}} & FedAvg  & 85.4 (11.2) & 10.0 (10.1) & 10.0 (10.1) \\ \cline{2-5} 
  & Trimmed-mean  & 84.9 (11.0) & 56.3 (16.0) & 20.5 (13.2) \\ \cline{2-5} 
  & Multi-krum  & 84.7 (11.3) & 58.8 (15.8) & 25.6 (14.4) \\ \cline{2-5} 
  & SignSGD  & 79.1 (12.8) & 39.7 (15.9) & 10.0 (10.1) \\ \cline{2-5} 
  & \textbf{\Name{}}  & \textbf{85.3 (11.3)} & \textbf{79.0 (12.4)} & \textbf{69.5 (14.8)} \\ \cline{2-5} 
  & \textbf{S\Name{} Top 50\% } & \textbf{77.6 (13.0)} & \textbf{41.7 (15.4)} & \textbf{39.7 (15.2)} \\ 
\hline \hline

  \multirow{6}{*}{{\vtop{\hbox{\strut FEMNIST + LeNet}\hbox{\strut 3400 clients}}}} & FedAvg  & 85.8 (10.2) & 6.3 (5.8) & 6.3 (5.8) \\ \cline{2-5} 
  & Trimmed-mean  & 85.2 (11.0) & 72.7 (15.7) & 56.2 (20.3) \\ \cline{2-5} 
  & Multi-krum  & 85.2 (10.9) & 80.9 (12.2) & 23.7 (12.8) \\ \cline{2-5} 
  & SignSGD  & 79.3 (12.4) & 76.7 (13.2) & 55.1 (14.9)  \\ \cline{2-5} 
  & \textbf{\Name{}}  & \textbf{84.2 (10.7)} & \textbf{83.0 (10.9)} & \textbf{65.8 (17.8)} \\ \cline{2-5} 
  & \textbf{S\Name{} Top 50\%}  & \textbf{75.2 (12.7)} & \textbf{70.5 (14.4)} & \textbf{60.39 (14.8)} \\ 
\hline
\end{tabular}
\end{table*}

\subsection{Experiment Setup}\label{sec:setupp}
Here, we introduce the datasets, the model architectures, the hyper-parameter settings, and FL baselines in more details.

\subsubsection{Datasets and model architectures} \label{app:dataset}
We use three benchmark datasets widely used in prior works on federated learning robustness:

\paragraphb{MNIST}  is a 10-class class-balanced classification task with 70,000 gray-scale images, each of size 28 × 28. We experiment with LeNet architecture given in Table~\ref{tab:models}. For local training in each \Name{}/FL round, each client uses 2 epochs. 
For training weights (experiments with FedAvg, SignSGD, TopK), we use SGD optimizer with learning rate of 0.01, momentum of 0.9, weight decay of 1e-4, and batch size 8. For training ranks (experiments with \Name{}), we use SGD with learning rate of 0.4, momentum 0.9, weight decay 1e-4, and batch size 8.

\paragraphb{CIFAR10}~\cite{krizhevsky2009learning} is a 10-class classification task with 60,000 RGB images (50,000 for training and 10,000 for testing), each of size 32 × 32. We experiment with a VGG-like architecture given in Table~\ref{tab:models}, which is identical to what \cite{ramanujan2020what} used.
For local training in each \Name{}/FL round, each client uses 5 epochs.
For training weights (experiments with FedAvg, SignSGD, TopK), we use SGD with learning rate of 0.1, momentum  of 0.9, weight decay of 1e-4, and batch size of 8. For training ranks (experiments with \Name{}), we optimize SGD with learning rate of 0.4, momentum of 0.9, weight decay of 1e-4, and batch size of 8. 

\paragraphb{FEMNIST}~\cite{DBLP:journals/corr/abs-1812-01097, DBLP:conf/ijcnn/CohenATS17}
is a character recognition classification task with 3,400 clients, 62 classes (52 for upper and lower case letters and 10 for digits), and 671,585 gray-scale images. Each client has data of their own handwritten digits or letters. We use LeNet architecture given in Table~\ref{tab:models}. 
For local training in each \Name{}/FL round, each client uses 2 epochs. 
For training weights (experiments with FedAvg, SignSGD, TopK), we use SGD with learning rate of 0.15, momentum of 0.9, weight decay of 1e-4,  and batch size of 10. For training ranks (experiments with \Name{}), we optimize SGD with learning rate of 0.2, momentum of 0.9, weight decay of 1e-4, and batch size of 10.


We optimize the hyperparameters based on \Name{} and other baselines independently.
In Appendix~\ref{sec:hyper-tune}, we show that robustness of \Name{} still persists even if we change the hyperparameters.

\subsubsection{Baseline FL Algorithms}\label{exp_setup:baseline_fl}
We compare the \Name{} with following FL baselines:

\paragraphb{Federated averaging:} 
In non-adversarial FL settings, i.e., without any malicious clients, the dimension-wise Average (FedAvg)~\cite{konevcny2016federated, mcmahan2017communication}  is an effective AGR. In fact, due to its efficiency, Average is the only AGR implemented by FL applications in practice~\cite{DBLP:journals/corr/abs-2007-10987, DBLP:journals/corr/abs-2102-08503}. 

\begin{table*}[ht]
\caption{Comparing the utility (test accuracy) and communication cost of FedAvg, \Name{} (in \textbf{bold}), SignSGD and, TopK and Sparse-\Name{} (S\Name{}) with different percentages of sparsity (in \textbf{bold}).} \label{tab:comm}
\centering
\begin{tabular} {|c|c||c|c|c|}
  \hline
  Dataset & Algorithm &  Accuracy (STD) & Upload (MB) & Download (MB) \\ \cline{1-5}
  \multirow{7}{*}{{\vtop{\hbox{\strut MNIST + LeNet}\hbox{\strut 1000 clients}}}} & FedAvg & 98.8 (3.1) & 6.20  & 6.20 \\ \cline{2-5}
  & \textbf{\Name{}} &  \textbf{98.8 (3.2)}  & \textbf{4.05} &  \textbf{4.05}  \\ \cline{2-5} 
  & \textbf{S\Name{} Top 50\%} &  \textbf{98.2 (3.8)}  & \textbf{2.03} &  \textbf{4.05}  \\ \cline{2-5} 
  & \textbf{S\Name{} Top 10\%} &  \textbf{89.5 (9.2)}  & \textbf{0.40} &  \textbf{4.05}  \\ \cline{2-5} 
  & SignSGD &  97.2 (4.6)  & 0.19 &  6.20  \\ \cline{2-5} 
  & TopK 50\% &  98.8 (3.2)  & 3.29 &  6.20  \\ \cline{2-5} 
  & TopK 10\% &  98.7 (3.2)  & 0.81 &  6.20  \\ \cline{2-5}  
\hline \hline
  \multirow{7}{*}{{\vtop{\hbox{\strut CIFAR10 + Conv8}\hbox{\strut 1000 clients}}}} & FedAvg & 85.4 (11.2) & 20.1 & 20.1\\ \cline{2-5}
  & \textbf{\Name{}} &  \textbf{85.3 (11.3)}  & \textbf{13.1} &  \textbf{13.1}  \\ \cline{2-5} 
  & \textbf{S\Name{} Top 50\%} & \textbf{77.6 (13.0)}   & \textbf{6.5} & \textbf{13.1}  \\ \cline{2-5} 
  & \textbf{S\Name{} Top 10\%} &  \textbf{27.5 (14.4)}  & \textbf{1.3} &  \textbf{13.1}  \\ \cline{2-5} 
  & SignSGD &  79.1 (13.6)  & 0.63 &  20.1  \\ \cline{2-5} 
  & TopK 50\% &  82.1 (11.8)  & 10.69 &  20.1 \\ \cline{2-5} 
  & TopK 10\% &  77.8 (13.0)  & 2.64 &  20.1  \\ \cline{2-5} 
\hline \hline
  \multirow{7}{*}{{\vtop{\hbox{\strut FEMNIST + LeNet}\hbox{\strut 3400 clients}}}} & FedAvg & 85.8 (10.2) & 6.23 & 6.23\\ \cline{2-5}
  & \textbf{\Name{}} &  \textbf{84.2 (10.7)}  & \textbf{4.06} &  \textbf{4.06}  \\ \cline{2-5} 
  & \textbf{S\Name{} Top 50\%} &  \textbf{75.2 (12.7)}  & \textbf{2.03} &  \textbf{4.06}  \\ \cline{2-5} 
  & \textbf{S\Name{} Top 10\%} &  \textbf{59.2 (15.0)}  & \textbf{0.40} &  \textbf{4.06}  \\ \cline{2-5} 
  & SignSGD &  79.3 (12.4)  & 0.19 &  6.23  \\ \cline{2-5} 
  & TopK 50\% &  85.7 (9.9)  & 3.31 &  6.23  \\ \cline{2-5} 
  & TopK 10\% & 85.5 (10.0) & 0.81 &  6.23  \\ \cline{2-5} 
\hline
\end{tabular}
\end{table*}

\paragraphb{SignSGD:} is a quantization method used in distributed learning to compress each dimension of gradient updates into 1 bit instead of 32 or 64 bits. To achieve this, in SignSGD~\cite{bernstein2018signsgd} the clients only send the sign of their gradient updates to the server, and the server runs a majority vote on them. 
SignSGD is designed for distributed learning where all the clients participate in each round, so all the clients are aware of the most updated weight parameters of the global model. However, SignSGD only reduces upload communication (clients$\rightarrow$server). But, does not reduce download communication (server$\rightarrow$clients), i.e., to achieve good performance of the global model,  the server sends all the weight parameters (each of 32 bits) to the newly selected clients in each round.
Hence, SignSGD 
is as inefficient as FedAvg in download communication.

\paragraphb{TopK:} is a sparsification method used in distributed learning that transmits only a few dimensions of each model update to the server. 
In TopK~\cite{DBLP:conf/emnlp/AjiH17,AlistarhH0KKR18}, the clients first sort the absolute values of their local model updates, and send the Top K\% largest model update dimensions to the server for aggregation. 
TopK suffers from the same problem as SignSGD: for performance reasons, the server should send the entire updated model weights to the new selected clients.


\paragraphb{Multi-krum:} \cite{blanchard2017machine} proposed Multi-krum
AGR as a modification to their own Krum AGR. Multi-krum selects an update using Krum and adds it to a selection
set, $S$. Multi-krum repeats this for the remaining updates (which remain after removing the update that Krum selects) until $S$ has $c$ updates such that $n-c>2m+2$, where $n$ is the number of selected clients and $m$ is the number of compromised clients in a given round. Finally, Multi-krum averages the updates in $S$.

\paragraphb{Trimmed-mean:} Yin et al.~\cite{YinCRB18} proposed Trimmed-mean that aggregates
each dimension of input updates separately. It sorts the values of the $j^{th}$-dimension of all updates. Then it removes $m$ (i.e., the number of compromised clients) of the largest and smallest values of that dimension, and computes the average of the rest of the values as its aggregate for the dimension $j$.

\subsection{Security Analysis}\label{Exp:security}

\begin{table*}[ht]
\centering
\caption{Comparing the performance of \Name{} and FedAvg in cross-device FL setting using two non-iid data distribution methods. We distribute data among 1000 clients with two methods described briefly below; please check Section~\ref{sec:noniid2} for more details.
} \label{T:Non}
\vspace*{0.25em}
\setlength{\extrarowheight}{0.02cm}
\begin{tabular} {|c|c|c||c||c||}
  \hline
  \multirow{2}{*}{Dataset} & \multirow{2}{*}{Type of Non-IID} & \multirow{2}{*}{Metric} & \multicolumn{2}{c|}{Algorithm} \\ \cline{4-5}
  
  & & & FedAvg & \Name{}\\ \cline{1-5}
  \multirow{8}{*}{\vtop{\hbox{\strut MNIST}\hbox{{\vtop{\hbox{\strut LeNet}\hbox{\strut N=1000}}}}}} &\multirow{4}{*}{Dirichlet Distribution $\beta=1$} & Mean & 98.8 & 98.8 \\ \cline{3-5} 
  & & STD  & 3.1 & 3.1 \\ \cline{3-5} 
  & & Min & 75.0 & 75.0\\ \cline{3-5} 
  & & Max & 100 & 100\\ \cline{2-5}
  &\multirow{4}{*}{Randomly 2 classes assigned to each client} & Mean & 98.4  & 98.3 \\ \cline{3-5} 
  & & STD  & 4.3 & 4.1 \\ \cline{3-5} 
  & & Min & 70.0 & 80.0\\ \cline{3-5} 
  & & Max & 100 & 100 \\ 
\hline \hline
  \multirow{8}{*}{\vtop{\hbox{\strut CIFAR10}\hbox{{\vtop{\hbox{\strut Conv8}\hbox{\strut N=1000}}}}}} &\multirow{4}{*}{Dirichlet Distribution $\beta=1$} & Mean & 85.4 & 85.3 \\ \cline{3-5} 
  & & STD  & 11.2 & 11.3 \\ \cline{3-5} 
  & & Min & 33.3 & 33.3 \\ \cline{3-5} 
  & & Max & 100 & 100\\ \cline{2-5}
  &\multirow{4}{*}{Randomly 2 classes assigned to each client} & Mean & 70.6 & 70.9 \\ \cline{3-5} 
  & & STD  & 21.9 & 19.2 \\ \cline{3-5} 
  & & Min & 0 & 10.0\\ \cline{3-5} 
  & & Max & 100 & 100\\ 
\hline \hline
\end{tabular}
\end{table*}

We compare \Name{} with  state-of-the-art robust aggregation rules (AGRs): Mkrum~\cite{blanchard2017machine}, and Trimmed-mean~\cite{YinCRB18}. 
Table~\ref{tab:mal} gives the performances of robust AGRs, SignSGD, and \Name{} with different percentages of malicious clients using attacks proposed by  Shejwalkar et al.~\cite{shejwalkar2021manipulating}, Bernstein et al.~\cite{bernstein2018signsgd}, and Algorithm~\ref{alg:poison} respectively. Here, 
we make a rather impractical assumption in favor of the \emph{previous} robust AGRs: 
we assume that the server knows the exact \% of malicious clients in each FL round.  Note that, \emph{\Name{} does not require this knowledge}.

\paragraphb{\Name{} achieves higher robustness than state-of-the-art robust AGRs:} 
We note from Table~\ref{tab:mal} that, \Name{} is more robust to the presence of malicious clients who mount untargeted poisoning attacks, compared to Multi-Krum and Trimmed-mean, when percentages of malicious clients are 10\% and 20\%. For instance,  on CIFAR10, 10\% malicious clients can decrease the accuracy of FL models to 56.3\% and 58.8\% for Trimmed-mean and Multi-Krum respectively; 20\% malicious clients can decrease the accuracy of the FL models to 20.5\% and 25.6\% for Trimmed-mean and Multi-Krum respectively. On the other hand, \Name{} performance decreases to 79.0\% and 69.5\% for 10\% and 20\% attacking ratio respectively.

We make similar observations for MNIST and FEMNIST datasets: for FEMNIST, 10\% (20\%) malicious clients reduce accuracy of the global model from 85.8\% to 72.7\% (56.2\%) for Trimmed-Mean, and to 80.9\% (23.7\%) for Multi-krum, while \Name{} accuracy decreases to 83.0\% (65.8\%).

\paragraphb{\Name{} is more accurate than SignSGD:} 
First, we note that, in the absence of malicious clients, \Name{} is significantly more accurate than SignSGD. For instance, on CIFAR10 distributed in non-iid fashion among 1000 clients, \Name{} achieves 85.3\% while SignSGD achieves 79.1\% , or on FEMNIST, \Name{} achieves 84.2\% while SignSGD achieves 79.3\%. This is because, \Name{} clients send more nuanced information via rankings of their subnetworks compared to SignSGD, where clients just send the signs of their model updates. 

\paragraphb{\Name{} is more robust than SignSGD:} 
Next, we note from Table~\ref{tab:mal} that, \Name{} is more robust against untargeted poisoning attacks compared to SignSGD when percentages of malicious clients are 10\% and 20\%. For instance,  on CIFAR10, 10\% (20\%) malicious clients can decrease the accuracy of SignSGD model to 39.8\% (10.0\%). On the other hand, \Name{} performance decreases to 79.0\% and 69.5\% for 10\% and 20\% attacking ratio respectively. 
We make similar observations for MNIST and FEMNIST datasets: for FEMNIST, 10\% (20\%) malicious clients reduce accuracy of the global model from 85.8\% to 76.7\% (55.1\%) for SignSGD, while \Name{} accuracy decreases to 83.0\% (65.8\%).

\paragraphb{Sparse-\Name{} robustness:} We evaluate robustness of S\Name{} Top 50\% against 10\% and 20\% malicious clients. As we can see from~\ref{tab:mal}, by sending only top half of the local rankings, the accuracy goes from 85.3\% (\Name{}) to 77.6\% (S\Name{}). S\Name{} also can provide robustness to some extend, but adversary has more influence on the global ranking since half of the rankings are missing. For instance, on CIFAR10, 10\% (20\%) malicious clients can decrease the accuracy of global ranking to 41.7\% (39.7\%) from 77.6\%. Also for FEMNIST, 10\% (20\%) malicious clients can decrease the accuracy of global ranking to 70.5\% (60.39\%) from 75.2\%.
We can see when malicious clients' percentages are higher, S\Name{} can perform better compared to existing robust AGR.

\paragraphb{\Name{ versus FedAvg and TopK}}
We omit evaluating FedAvg and TopK, because even a single malicious client~\cite{blanchard2017machine} can jeopardize their performances.


\subsection{Communication Cost Analysis}\label{Exp:comp}

In \Name{}, both clients and server communicate just the edge ranks instead of weight parameters. Thus, \Name{} reduces both upload and download communication cost.
Table~\ref{tab:comm} illustrates the utility, i.e., the mean and standard deviation of all clients' test accuracies and, communication cost of \Name{} and state-of-the-art quantization (i.e., SignSGD~\cite{bernstein2018signsgd}) and sparsification (i.e., TopK~\cite{AlistarhH0KKR18,DBLP:conf/emnlp/AjiH17}) communication-reduction methods. 

\paragraphb{\Name{} versus SignSGD:} 
SignSGD in FL reduces only the upload communication, but for efficiency reasons, the server sends all of the weight parameters (each of 32 bits) to the newly selected clients. Hence, SignSGD has very efficient upload communication, but very inefficient download communication. For instance, on CIFAR10,  for both upload and download, \Name{} achieves 13.1MB each while SignSGD achieves 0.63MB and 20.1MB, respectively.

\paragraphb{\Name{} versus TopK:}
We compare \Name{} and TopK where $K\in\{10,50\}\%$. 
\Name{} is more accurate than Topk for MNIST and CIFAR10: on CIFAR10, \Name{} accuracy is 85.3\%, while TopK accuracies are 82.1\% and 77.8\% with $K$=50\% and $K$=10\%, respectively. Similar to SignSGD, Topk more efficiently reduces upload communication, but has very high download communication.
Therefore, the combined upload and download communication cost per client per round is 26.2MB for \Name{} and 30.79MB for TopK with $K$=50\%; note that, even then TopK performs worse than \Name{}.

\paragraphb{Communication cost reduction due to Sparse-\Name{} (S\Name{}):}
We now evaluate S\Name{} explained in Section~\ref{sec: SFSL}. 
In S\Name{} with top 50\% ranks, clients send the top 50\% of their ranks to the server, which reduces the upload bandwidth consumption by half. Please note that the download cost of S\Name{} is the same as \Name{} since the \Name{} server should send all the global rankings to the selected clients in each round. We note from Table~\ref{tab:comm} that, by sending fewer ranks, S\Name{} reduces upload communication at a small cost of performance. For instance, on CIFAR10, SF\Name{} with top 50\% reduces the upload communication by 50\% at the cost reducing accuracy from 85.4\% to 77.6\%.


\subsection{Performances of \Name{} with Different Heterogeneous Data Distribution Methods}\label{sec:noniid2}

So far, we evaluated all of our experiments when the data is distributed non-iid using Dirichlet distribution with parameter $\beta=1$. In this method of non-iid data distribution, all clients will get at least a few samples from each data class with non-zero probabilities that Dirichlet distribution generates. However, this non-iid data distribution need not represent all the practical FL settings. In fact, there may exist non-iid distributions that make training FL models more difficult.
Therefore in this section, we consider a more difficult setting where the data distribution is more non-iid. 

\paragraphb{Assigning only two classes to each client:}
We experiment with the more extreme non-iid distribution considered by McMahan et al.~\cite{mcmahan2017communication}. More specifically, to distribute of MNIST and CIFAR10 data among 1000 clients, we sort all the (i.e., combined train and test) MNIST and CIFAR10 data according to their classes and then we partition them into 2000 shards. Hence, each shards of training MNIST  has 30 images and each CIFAR10 shard has 25 images of a single class. Then we assign two random shards to each client resulting in each client having data from at most two classes. Therefore, in CIFAR10 experiments, each client has 50 training images, and 10 test images, and in MNIST experiments, each client has 60 training images and 10 test images. We only use this assignment in Section~\ref{sec:noniid2}.

Table~\ref{T:Non} shows the performances of \Name{} and FedAvg using different methods of non-iid assignment. We distribute the data between 1000 clients using:  (I) Dirichlet distribution with $\beta=1$ similar to~\cite{reddi2020adaptive, hsu2019measuring} and (II) the method described above from~\cite{mcmahan2017communication}. In Table~\ref{T:Non}, we note that \emph{\Name{} achieves similar performances as FedAvg for different heterogeneous data distribution methods}. For instance, on CIFAR10, FedAvg and \Name{} achieves similar performances of 85.4\% and 85.3\% respectively when data is distributed according to (I). Similarly, when data is distributed according to (II), FedAvg and \Name{} achieve similar performances of 70.6\% and 70.9\%, respectively. 

We make similar observations for MNIST as well: FedAvg achieves 98.8\% and 98.4\% for the two methods of data distribution respectively, while \Name{} achieves 98.8\% and 98.3\% accuracy. 




\begin{figure*}[ht]
\centering
\begin{subfigmatrix}{3}
\centering
  \subfigure[MNIST]{
  \includegraphics[width=0.65\columnwidth]{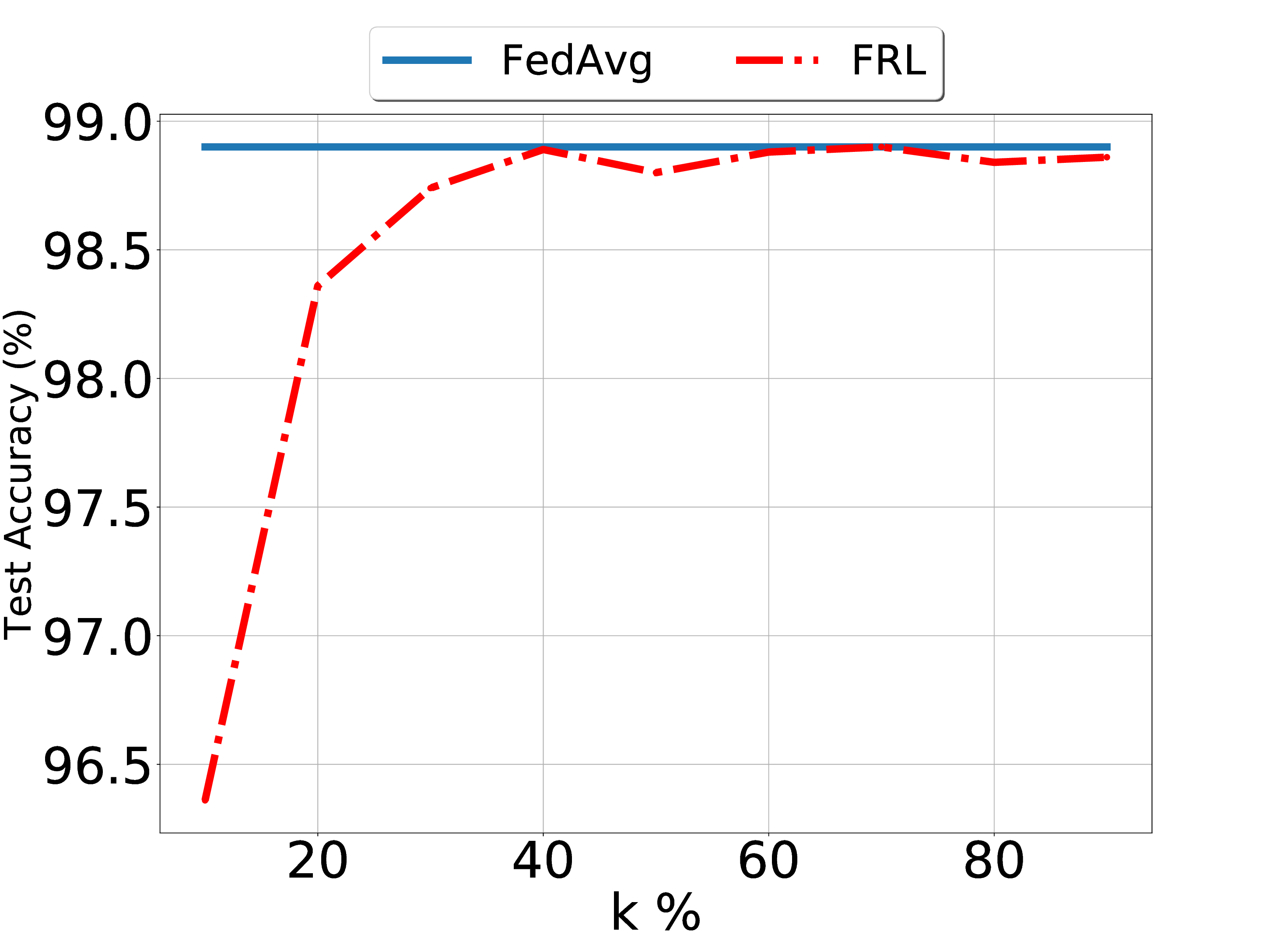}}
  \subfigure[CIFAR10]{
  \includegraphics[width=0.65\columnwidth]{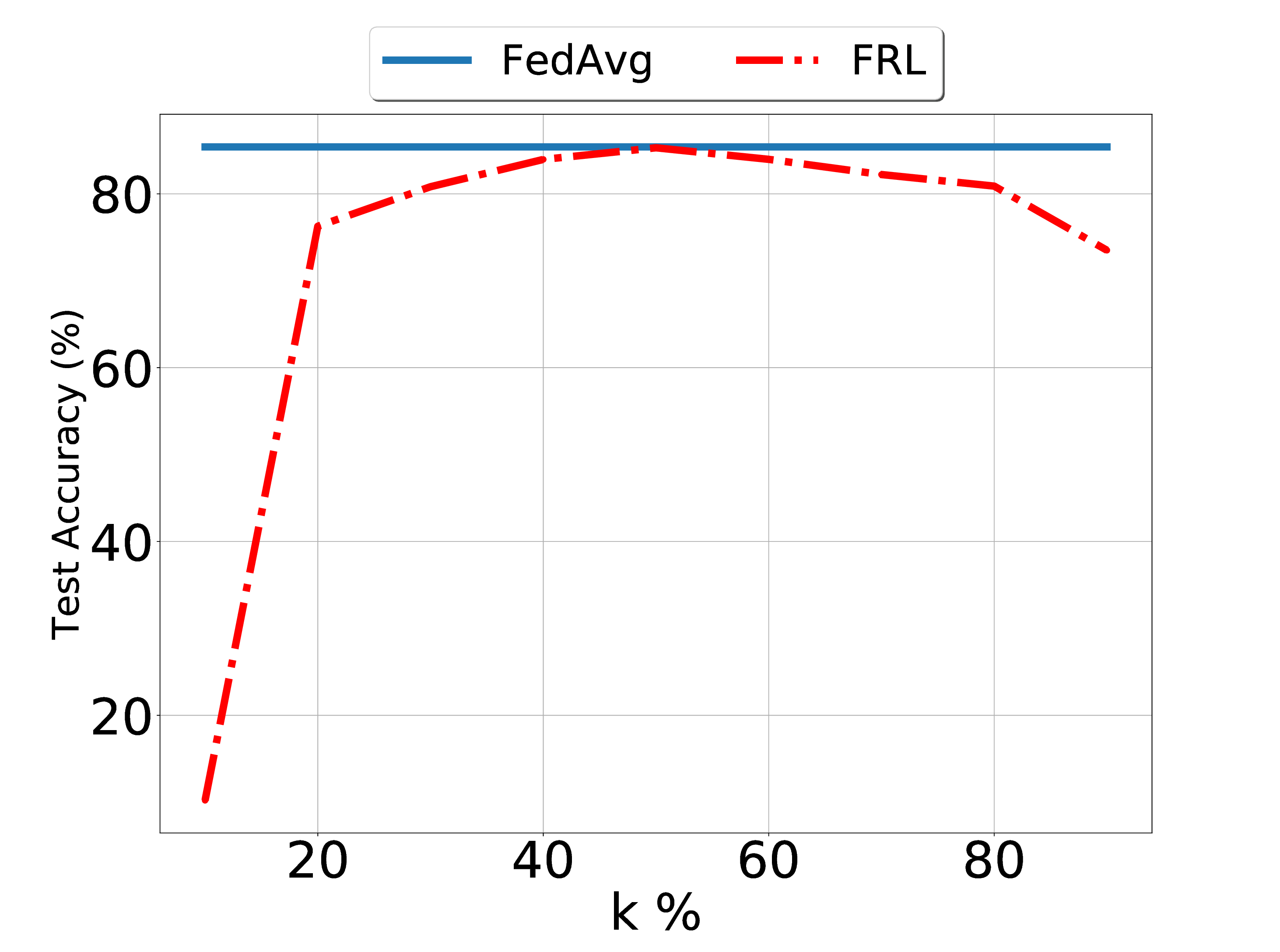}}
  \subfigure[FEMNIST]{
  \includegraphics[width=0.65\columnwidth]{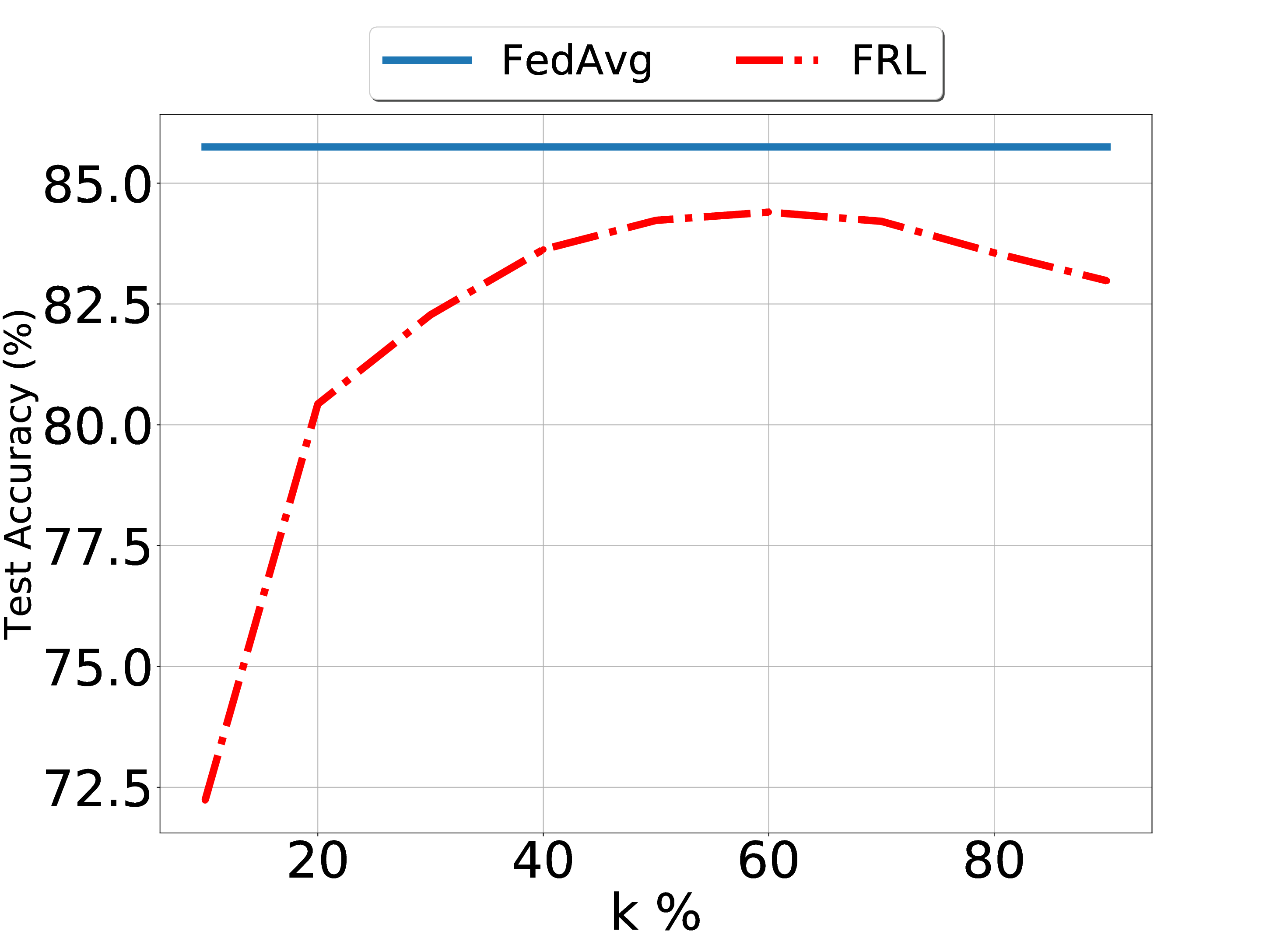}}
\end{subfigmatrix}    
\vspace{-10pt}
\caption{Comparing performance of \Name{} for different subnetwork sizes. $k$ (x-axis) shows the $\%$ of weights that each client is including in its subnetwork, test accuracy (y-axis) shows the mean of accuracies for all the clients on their test data. The chosen clients in each round send all the ranks to the server. \Name{} with subnetworks of $\in[40\%,70\%]$ result in better performances.}
\label{fig:spars}
\end{figure*}


\subsection{\Name{}: Weights Initialization Matters}\label{Exp:init}

\begin{table}[H]
\centering
\vspace{-10pt}
\caption{Comparing the performance of \Name{} with different random weight initialization algorithms with the performance of vanilla FedAvg for cross-device setting. Using Singed Kaiming Constant ($U_K$) as weight initialization gives the best performance for all the datasets. } \label{tab:init}
\setlength{\extrarowheight}{0.02cm}
\begin{tabular} {|c|c||c||c|c|c||}
  \hline
  \multirow{3}{*}{Dataset} & \multirow{2}{*}{Metric} & \multicolumn{4}{c|}{Algorithm} \\ \cline{3-6}
  
  & & FedAvg & \multicolumn{3}{c|}{\Name{}}\\ \cline{2-6}
    
     & $W_{init}\sim$ & - & $\mathcal{X}_N$ & $\mathcal{N}_K$ & $U_K$ \\ \hline \hline
  \multirow{4}{*}{\vtop{\hbox{\strut MNIST}\hbox{{\vtop{\hbox{\strut LeNet}\hbox{\strut N=1000}}}}}} & Mean & 98.8 & 96.6 & 98.7 & 98.8  \\ \cline{2-6} 
  & STD  & 3.1 & 5.2 & 3.2 & 3.1 \\ \cline{2-6} 
  & Min & 75.0 & 57.1 & 75.0 & 75.0 \\ \cline{2-6} 
  & Max & 100 & 100 & 100 & 100\\ 
\hline \hline
  \multirow{4}{*}{\vtop{\hbox{\strut CIFAR10}\hbox{{\vtop{\hbox{\strut Conv8}\hbox{\strut N=1000}}}}}} &  Mean& 85.4 & 63.6 & 82.0 &85.3  \\ \cline{2-6} 
  & STD & 11.2 & 15.6 & 11.9 & 11.3 \\ \cline{2-6} 
  & Min  & 33.3 & 0 & 0 & 33.3 \\ \cline{2-6} 
  & Max  & 100 & 100 & 100 &100  \\ 
  \hline \hline
    \multirow{4}{*}{\vtop{\hbox{\strut FEMNIST}\hbox{{\vtop{\hbox{\strut LeNet}\hbox{\strut N=3400}}}}}} &  Mean & 85.8 & 69.2 &  82.9 &84.2  \\ \cline{2-6} 
  & STD & 10.2 & 14.2 &  11.1 & 10.7  \\ \cline{2-6} 
  & Min & 10.0 & 0 & 14.3 & 7.1  \\ \cline{2-6} 
  & Max & 100 & 100 &  100 &100  \\ 
\hline \hline
\end{tabular}
\end{table}

In \Name{}, the weight parameters are randomly initialized at the start and remain fixed throughout the training. An appropriate initialization is instrumental to the success of \Name{}, since the FRL clients are sending the local rankings of these edges; more important edges get higher ranks. They generate these rankings by feeding the subnetwork of top rank edges and calculating the gradient of the loss with respect to the scores, so distribution of these random weights has a high impact on the calculated loss.
We use three different distribution for initializing the weight parameters as follows:



\paragraphb{Glorot Normal}~\cite{DBLP:journals/jmlr/GlorotB10} where we denote by $\mathcal{X}_N$. Previous work~\cite{DBLP:conf/nips/ZhouLLY19} used this initialization to demonstrate that subnetworks of randomly weighted neural networks can achieve impressive performance. 

\paragraphb{Kaiming Normal}~\cite{he2015delving} where we denote by $\mathcal{N}_k$ defined as $\mathcal{N}_K = \mathcal{N} \left(0, \sqrt{2/n_{\ell-1}} \right)$ where $\mathcal{N}$ shows normal distribution. $n_{\ell}$ shows the number of parameters in the $\ell$th layer.

\paragraphb{Singed Kaiming Constant}~\cite{ramanujan2020what} where all the wights are a constant $\sigma$ but they are assigned $\{+,-\}$ randomly. This constant, $\sigma$, is the standard deviation of Kaiming Constant. We show this initialization with $U_K$ as we are sampling from
$\{-\sigma, +\sigma\}$ where $\sigma=\left(\sqrt{2/n_{\ell-1}} \right)$.

Table~\ref{tab:init} shows the results of running \Name{} for three datasets under the three aforementioned initialization algorithms. We compare \Name{} with FedAvg and report the mean, standard deviation, minimum, and maximum of the accuracies for the clients' accuracies in  \Name{} and FedAvg at the end of training. As we can see under three different random initialization, using Signed Kaiming Constant ($U_K$) results in better performance. 
We note from Table~\ref{tab:init} that \Name{} with Signed Kaiming Constant ($U_K$) initialization achieves performance very close to the performance of FedAVg. 

Note that, since the \Name{} clients update scores in each round, unlike initialization of weights, initialization of scores does not have significant impact on the final global subnetwork search. Therefore, we do not explore different randomized initialization algorithms for scores and simply use Kaiming Uniform initialization for scores.


Ramanujan et al.~\cite{ramanujan2020what} also considered these three initialization to find the best subnetwork in centralized machine learning setting. They also showed that using Singed Kaiming Constant gives the best supermasks. Our results align with their conclusions, hence we use Singed Kaiming Constant to initialize the weights and Kaiming Uniform to initialize the scores of the global supernetwork.

\subsection{Performances of \Name{} with Varying Sizes of Subnetworks}\label{sec:size}
In \Name{}, each client uses Edge-popup (Algorithm~\ref{alg:edgepop}) and their local data to find a local ranking by finding a subnetwork within a randomly initialized global network, which we call \emph{supernetwork}. Edge-popup algorithm uses parameter $k$ which represents the $\%$ of all the  edges in a supernetwork which will remain in the final subnetwork.
For instance, $k=50\%$ denotes that each client finds a subnetwork within a supernetwork that has half the number of edges as in the supernetwork.

Figure~\ref{fig:spars} illustrates how the performance of \Name{} varies with the sizes of local subnetworks that the clients share with the server. In other words, when we vary the sparsity $k\%$ of edge popup algorithm during local subnetwork search $k \in [10, 20, 30, 40, 50, 60, 70, 80, 90]\%$.
Interestingly we note that, \Name{} performs the worst  when clients use all ($k$=100\%) or none ($k$=0\%) of the edges. This is because, it is difficult to find a subnetwork with small number of edges. While using all of the edge essentially results in using a random neural network.
As we can see \Name{} with $k \in [40,70]\%$, gives the best performances for all the three datasets. Hence, we set $k$=50\% by default in our experiments.



Due to space limitations, we defer more experiments of \Name{} to Appendix~\ref{App:missing1}.

%



\section{Conclusions}\label{conclusion}

We designed a novel collaborative learning algorithm, called \emph{\textbf{F}ederated \textbf{R}ank \textbf{L}earning} (\Name{}), to address the issues of robustness to poisoning and communication efficiency in existing FL algorithms.
We argue that a core reason for the susceptibility of existing FL algorithms to poisoning is the large continuous space of values in their model updates.
Hence, in \Name{}, we use ranks of edges of a randomly initialized neural network contributed by collaborating clients to find a global ranking and then use a subnetwork based only on the top edges.
Use of rankings in a fixed range restricts the space available to poisoning adversaries to craft malicious updates, and also allows \Name{} to reduce the communication cost.
%
We show, both theoretically and empirically, that ranking based collaborative learning can effectively mitigate the robustness issue as well as reduce the communication costs involved.

\section*{Acknowledgements}
This work was supported by NSF grant 2131910. 

\newpage


\bibliographystyle{plain}
\bibliography{main.bib}

\appendix

\section{Theoretical analysis of robustness of \Name{}} \label{app:Robust1}

In this section, we detail the proof of robustness of \Name{}. 
In other words, we prove an upper bound on the failure probability of robustness of \Name{}, i.e., the probability that a good edge will be removed from the final subnetwork when malicious clients mount the worst case attack.
{Inspired from SignSGD~\cite{bernstein2018signsgd}, for this proof, We assume a simpler \textsc{Vote(.)} function where if more than half of the clients add an edge $e_i$ to their subnetworks, then the \Name{} server adds it to the final global subnetwork. We also assume that the malicious clients cannot collude in our proof. 
} 

Assume that edge $e_i$ is a good edge, i.e., having $e_i$ in the final subnetwork improves the performance of the  final subnetwork. 
Let Z be the random variable that represents the number of clients who vote for the edge $e_i$ to be in the final subnetwork, i.e., the number of clients whose local subnetwork of size $k$\% of the entire supernetwork (Algorithm~\ref{alg:FSL+} line 11) contains $e_i$.
Therefore, $Z \in [0,n]$ where $n$ is the number of clients being processed in a given \Name{} round.

Let G and B be the random variable that represent the number of benign and malicious clients that vote for edge $e_i$, respectively; the malicious clients inadvertently exclude the good edge $e_i$  in their local subnetwork based on their benign training data.

There are total of $\alpha n$ malicious clients, where $\alpha$ is the fraction of malicious clients that B of them decides that $e_i$ is a bad edge and should not be removed. 
Each of the malicious clients computes the subnetwork on its own benign training data, so B of them do not conclude that $e_i$ is a good edge. Hence, $Z=G+B$. We can say that G and B have binomial distribution
, i.e., $G \sim \text{binomial}([(1-\alpha)n, p]$ and $B \sim \text{binomial}([\alpha n, 1-p]$ where $p$ is the probability that a benign client has this edge in their local subnetwork and $\alpha$ is the fraction of malicious clients. 
Note that the probability that our voting in simplified \Name{} fails is $P[\text{failure}] = P[Z <= \frac{n}{2}]$, i.e., when more than half of the clients vote against $e_i$, i.e., they do not include $e_i$ in their local subnetworks.
We can find the mean and variance of Z as follows:

\begin{equation}
E[Z]=(1-\alpha)np + \alpha n(1-p)
\end{equation}

\begin{equation}
Var[Z]=(1-\alpha)np(1-p)+\alpha np(1-p)=np(1-p)
\end{equation}

\cite{cantelli1929sui} provides an inequality where for a random variable X with mean $\mu$ and variance $\sigma^2$ we have $P[\mu -X >= \lambda] <= \frac{1}{1+\frac{\lambda^2}{\sigma^2}}$. Using this inequality, we can write:

\begin{equation}
    P[Z<=\frac{n}{2}]=P[E[Z]-Z>=E[Z]-n/2] <= \frac{1}{1+\frac{(E[z]-n/2)^2}{var[Z]}}
\end{equation}

because $1+x^2>=2x$, we have:
\begin{align}\label{eq:upper_bound}
    P[Z<=\frac{n}{2}]&<=1/2 \sqrt{\frac{Var[Z]}{(E[Z]-n/2)^2}} \\ \nonumber 
    &=1/2 \sqrt{\frac{np(1-p)}{(np-\alpha np +\alpha n -\alpha np - n/2)^2}} \\ \nonumber 
    &=1/2 \sqrt{\frac{np(1-p)}{(n(p +\alpha(1 -2 p) - 1/2))^2}}
\end{align}


What this means is that the probability that the simplified \textsc{Vote(.)} fails is upper bounded as in~\eqref{eq:upper_bound}.
We show the effect of the different values of $\alpha$ and $p$ in Figure~\ref{fig:upper}. We can see from Figure~\ref{fig:upper}, if the benign clients can train better supermasks (better chance that a good edge ended in their subnetwork), the probability that the attackers succeed is lower (more robustness).
\textsc{Vote(.)} in \Name{} (Section~\ref{fsl:design:vote}) is more sophisticated and puts more constraints on the malicious clients, hence the about upper bound also applies to \Name{}. 

\section{Missing information about \Name{} optimization function} \label{app:optimization}
Ramanujan et al.~\cite{ramanujan2020what} proved that when edge $(a,b)$ replaces $(c,b)$ in layer $\ell$ and the rest of the subnetwork remains fixed then the loss of the supermask learning decreases (provided the loss is sufficiently smooth).
Motivated by their proof, we show that when these two edges are swapped in \Name{}, the loss decreases for \Name{} optimization too.

\textbf{Theorem 1:} when edge $(a,b)$ replaces $(c,b)$ in layer $\ell$ and the rest of the subnetwork remains fixed then the loss of the \Name{} optimization will decrease (provided the loss is sufficiently smooth).

\textit{proof.}
First, we know that the optimization problem of \Name{} is as follow:

\begin{align}
    \min_{R_{g}} F(\theta^w, R_{g}) = \min_{R_{g}} \sum_{i=1}^N \lambda_i L_i(\theta^w \odot \mathbf{m}) \\ \nonumber 
    \; \; \text{s.t.}  \; \;\mathbf{m}[R_{g}<k]=0 \; \; \text{and} \; \; \mathbf{m}[R_{g}\geq k]=1 
\end{align}

where $\lambda_i$ shows the importance of the $i^{th}$ client in empirical risk minimization which $\lambda_i=\frac{1}{N}$ gives same importance to all the participating clients. $\mathbf{m}$ is the final mask that contains the edges of top $k$ ranks, and $L_i$ is the loss function for the $i$th client. $\theta^w \odot \mathbf{m}$ shows the subnetwork inside the random $\theta^w$ that all clients unanimously vote for. In this optimization, the \Name{} clients try to minimize $F$ by finding the best global ranking $R_{g}$.

We now wish to show $F(\theta^w, R_g^{t+1}) < F(\theta^w, R_g^{t})$ when in \Name{} round $t+1$, the edge $(a,b)$ replaces $(c,b)$ in layer $\ell$ and the rest of the subnetwork remains fixed.
Suppose global rank of edge $(a,b)$ was $R_{g}^t[(a,b)]$ and global rank of edge $(c,b)$ was $R_g^t[(c,b)]$ in round $t$, so we have:

\begin{align}
R_{g}^{t}[(a,b)]<R_{g}^t[(c,b)] \\
R_{g}^{t+1}[(a,b)]>R_{g}^{t+1}[(c,b)]
\end{align}
where the order of all the remaining global ranks remains fixed, and only these two edges are swapped in global ranking.
Now let $s_{ab}^{t,i}$ shows the score of weight $w_{ab}$
in round $t$ and $i^{th}$ client and $s_{ab}^{t+1,i}$ shows the updated score of it after local training. As in our majority vote, we are calculating the sum of the reputation of edges we will have:

\begin{align}\label{eq1}
\sum_{i=1}^N s_{ab}^{t,i} < \sum_{i=1}^N  s_{cb}^{t,i} 
\end{align}

\begin{align}\label{eq11}
\sum_{i=1}^N  s_{ab}^{t+1,i} > \sum_{i=1}^N  s_{cb}^{t+1,i}
\end{align}

We also know that Edge-popup algorithm updates the scores in the $i^{th}$ client as follow:
\begin{align}\label{eq2}
s_{ab}^{t+1, i} = s_{ab}^{t, i}- \eta \frac{\partial L}{\partial I_a} Z_a W_{ab}
\end{align}

Based on~\eqref{eq1}, and ~\eqref{eq11}, we can say:

\begin{align}\label{eq4}
\sum_{i=1}^N s_{ab}^{t,i} - \sum_{i=1}^N  s_{cb}^{t,i} < \sum_{i=1}^N s_{ab}^{t+1,i} - \sum_{i=1}^N  s_{cb}^{t+1,i} 
\end{align}

And based on~\eqref{eq2}, we also know that:

\begin{equation}\label{eq5}
    \sum_{i=1}^N \left( s_{ab}^{t+1, i} - s_{ab}^{t , i} \right)= \sum_{i=1}^N \left( - \eta \frac{\partial L^i}{\partial I_a^i} Z_a^i W_{ab} \right)
\end{equation}

\begin{equation}\label{eq6}
    \sum_{i=1}^N \left( s_{cb}^{t+1, i} - s_{cb}^{t , i} \right)= \sum_{i=1}^N \left( - \eta \frac{\partial L^i}{\partial I_c^i} Z_c^i W_{cb} \right)
\end{equation}

Based on~\eqref{eq4},~\eqref{eq5} and~\eqref{eq6}, we can say:

\begin{equation}\label{eq8}
    \sum_{i=1}^N \left( \frac{\partial L^i}{\partial I_c^i} Z_c^i W_{cb} \right) > \sum_{i=1}^N \left(\frac{\partial L^i}{\partial I_a^i} Z_a^i W_{ab} \right)
\end{equation}


So based on~\eqref{eq8}, and what~\cite{ramanujan2020what} proved for each supermask training we can show~\eqref{eq:10}. We assume that loss is smooth and the input to the nodes that their edges are swapped are close before and after the swap. 

\begin{equation}\label{eq:10}
    \sum_{i=1}^N \left( L_i(\theta^w \odot \mathbf{m^{t+1}}) \right) < \sum_{i=1}^N \left( L_i (\theta^w \odot \mathbf{m}^{t}) \right)
\end{equation}

that means:
\begin{equation}
    F(\theta^w, R_{g}^{t+1}) < F(\theta^w, R_{g}^{t})
\end{equation}

\section{Missing Details of Experimental Setup} \label{app:setup}

\subsection{Model Architectures}
Table~\ref{tab:models} show the model architectures and the number of parameters in each layer for them. 

\begin{table}[t]
\caption{Model architectures. We use identical architecture to those ~\cite{ramanujan2020what, wortsman2020supermasks} used. } \label{tab:models}
\centering
\fontsize{9}{10}\selectfont{}
\setlength{\extrarowheight}{0.05cm}
\begin{tabular} {|c|c|c|}
  \hline
  Architecture & Layer Name & Number of parameters \\ \hline \hline
 
 \multirow{5}{*}{{\vtop{\hbox{\strut LeNet~\cite{wortsman2020supermasks}}\hbox{\strut MNIST}}}} & Convolution(32) + Relu &  288 \\ \cline{2-3}
  & Convolution(64) + Relu &  18432 \\ \cline{2-3}
  & MaxPool(2x2) &  - \\ \cline{2-3}
 & FC(128) + Relu &  1605632 \\ \cline{2-3}
  & FC(10) &  1280 \\ \cline{2-3}
\hline \hline 

  \multirow{15}{*}{{\vtop{\hbox{\strut Conv8~\cite{ramanujan2020what}}\hbox{\strut CIFAR10}}}} & Convolution(64) + Relu &  1728 \\ \cline{2-3}
  & Convolution(64) + Relu &  36864 \\ \cline{2-3}
  & MaxPool(2x2) &  - \\ \cline{2-3}
  & Convolution(128) + Relu &  73728 \\ \cline{2-3}
  & Convolution(128) + Relu &  147456 \\ \cline{2-3}
  & MaxPool(2x2) &  - \\ \cline{2-3}
  & Convolution(256) + Relu &  294912 \\ \cline{2-3}
  & Convolution(256) + Relu &  589824 \\ \cline{2-3}
  & MaxPool(2x2) &  - \\ \cline{2-3}
  & Convolution(512) + Relu &  1179648 \\ \cline{2-3}
  & Convolution(512) + Relu &  2359296 \\ \cline{2-3}
  & MaxPool(2x2) &  - \\ \cline{2-3}
 & FC(256) + Relu &  524288 \\ \cline{2-3}
  & FC(256) + Relu &  65536 \\ \cline{2-3}
  & FC(10) &  2560 \\ \cline{2-3}
\hline \hline

  \multirow{5}{*}{{\vtop{\hbox{\strut LeNet~\cite{wortsman2020supermasks}}\hbox{\strut FEMNIST}}}} & Convolution(32) + Relu &  288 \\ \cline{2-3}
  & Convolution(64) + Relu &  18432 \\ \cline{2-3}
  & MaxPool(2x2) &  - \\ \cline{2-3}
 & FC(128) + Relu &  1605632 \\ \cline{2-3}
  & FC(62) &  7936 \\ \cline{2-3}
\hline 
\end{tabular}
\end{table}

\subsection{hyperparameters Tuning} \label{sec:hyper-tune}
We optimize the hyperparameters based on \Name{} and other baselines independently.
The hyperparameters that we used in our experiments are tuned in scenario with no malicious clients.  Table~\ref{T:hyper1} shows the performance of \Name{} and other baselines on CIFAR10 (distributed over 1000 users using Dirichlet distribution) for different values of hyperparameters when there are 10\% malicious clients among the clients. 
This table shows the robustness of \Name{} still persists even if we change the hyperparameters. 
We reported mean of accuracies and standard deviation of accuracies for all the clients at the final \Name{} round. 

\subsection{Model Poisoning Attack for Robustness Evaluations}\label{exp_setup:attack}
To evaluate robustness of various FL algorithms, we use state-of-the-art model poisoning attack proposed by ~\cite{shejwalkar2021manipulating} in our robustness experiments.
The attack proposes a general FL poisoning framework and then tailors it to specific FL settings. It first computes an average $\nabla^b$ of the available benign updates  and perturbs it in a \emph{dynamic, data-dependent malicious direction} $\omega$ to compute the final poisoned update $\nabla'= \nabla^b + \gamma \omega$. DYN-OPT finds the largest $\gamma$ that successfully circumvents the target AGR. 
DYN-OPT is much stronger, because unlike STAT-OPT, it finds the largest $\gamma$ and uses a dataset tailored $\omega$. Privacy preservation~\cite{mozaffari2020heterogeneous} is another major challenge to FL, but is orthogonal to our work.

\begin{table*} 
\caption{Performance of \Name{} with different hyperparameters trained on CIFAR10 (distributed over 1000 clients using Dirichlet distribution). } \label{T:hyper1}
\centering
\begin{tabular} {|c|c|c|c|}
  \hline
  Method & hyperparameter & value & Test Accuracy with 10\% malicious\\ \cline{1-4}
 \multirow{11}{*}{\Name{}} &  \multirow{3}{*}{batch size} & 6  & 78.4 (12.6) \\ \cline{3-4}
 & & {8} &   {79.0 (12.4)} \\ \cline{3-4}
 & & 16 & 76.4 (13.6)\\ \cline{2-4}
 &  \multirow{3}{*}{local epochs} & 2 & 79.8 (12.2) \\ \cline{3-4}
 & & {5}  & {79.0 (12.4)} \\ \cline{3-4}
 & & 10  & 78.2 (12.6)  \\ \cline{2-4}
 &  \multirow{5}{*}{learning rate} & 0.1 & 73.5 (13.4) \\ \cline{3-4}
 & & 0.2 & 82.4 (12.1) \\ \cline{3-4}
 & & 0.3  & 83.11 (11.8) \\ \cline{3-4}
 & & {0.4} & {79.0 (12.4)}\\ \cline{3-4}
 & & 0.5  & 77.5 (13.1)\\ \cline{2-4}

 \hline \hline 
 FedAvg & - & - & {10.0 (10.1)} \\ \cline{3-4}
 \hline \hline 
 TopK & - & - & {10.0 (10.1)} \\ \cline{3-4}
 
 \hline \hline 
 \multirow{11}{*}{FedAvg + Trimmed-mean} &  \multirow{3}{*}{batch size} & 6  & 55.5 (14.5)\\ \cline{3-4}
 & & {8} &   {56.3 (16.0)}\\ \cline{3-4}
 & & 16 & 37.7 (15.6) \\ \cline{2-4}
 &  \multirow{3}{*}{local epochs} & 2 & 41.0 (15.4)\\ \cline{3-4}
 & & {5}  & {56.3 (16.0)}\\ \cline{3-4}
 & & 10  & 21.0 (9.9)  \\ \cline{2-4}
 &  \multirow{5}{*}{learning rate} & 0.01 & 34.0 (15.5)\\ \cline{3-4}
 & & 0.05 & 38.3 (15.3) \\ \cline{3-4}
 & & 0.1  & {56.3 (16.0)} \\ \cline{3-4}
 & & 0.15  & 10.0 (10.0) \\ \cline{3-4}
 & & 0.2 & 10.0 (10.0)\\ \cline{3-4}
 
 \hline \hline 
 \multirow{11}{*}{FedAvg + Multi-Krum} &  \multirow{3}{*}{batch size} & 6  & 19.0 (12.5) \\ \cline{3-4}
 & & {8} &   {58.8 (15.8)}\\ \cline{3-4}
 & & 16 & 36.7 (14.8) \\ \cline{2-4}
 &  \multirow{3}{*}{local epochs} & 2 & 46.1 (15.9)\\ \cline{3-4}
 & & {5}  & {58.8 (15.8)}\\ \cline{3-4}
 & & 10  & 24.3 (11.7)  \\ \cline{2-4}
 &  \multirow{5}{*}{learning rate} & 0.01 & 15.3 (11.7) \\ \cline{3-4}
 & & 0.05 & 50.0 (16.2) \\ \cline{3-4}
 & & 0.1  & {58.8 (15.8)} \\ \cline{3-4}
 & & 0.15  & 15.4 (11.9)\\ \cline{3-4}
 & & 0.2 & 10.0 (10.0)\\ \cline{3-4}
 
 \hline \hline 
 \multirow{11}{*}{SignSGD} &  \multirow{3}{*}{batch size} & 6  & 33.1 (15.6)\\ \cline{3-4}
 & & {8} &   {39.7 (15.9)}\\ \cline{3-4}
 & & 16 & 10.2 (10.1) \\ \cline{2-4}
 &  \multirow{3}{*}{local epochs} & 2 & 10.2 (10.5)\\ \cline{3-4}
 & & {5}  & {39.7 (15.9)}\\ \cline{3-4}
 & & 10  & 41.5 (16.0)  \\ \cline{2-4}
 &  \multirow{5}{*}{learning rate} & 0.01 & 44.2 (15.8) \\ \cline{3-4}
 & & 0.05 & 41.9 (15.5) \\ \cline{3-4}
 & & 0.1  & {39.7 (15.9)} \\ \cline{3-4}
 & & 0.15  & 35.8 (15.3)\\ \cline{3-4}
 & & 0.2 & 10.2 (10.1)\\ \cline{3-4}
 
\hline 
\end{tabular}
\end{table*}

\section{Missing Experiments of \Name{}}\label{App:missing1}

\begin{figure}
\centering
\begin{subfigmatrix}{1}
\centering
  \subfigure[CIFAR10 (Test Accuracy)]{
  \includegraphics[width=1\columnwidth]{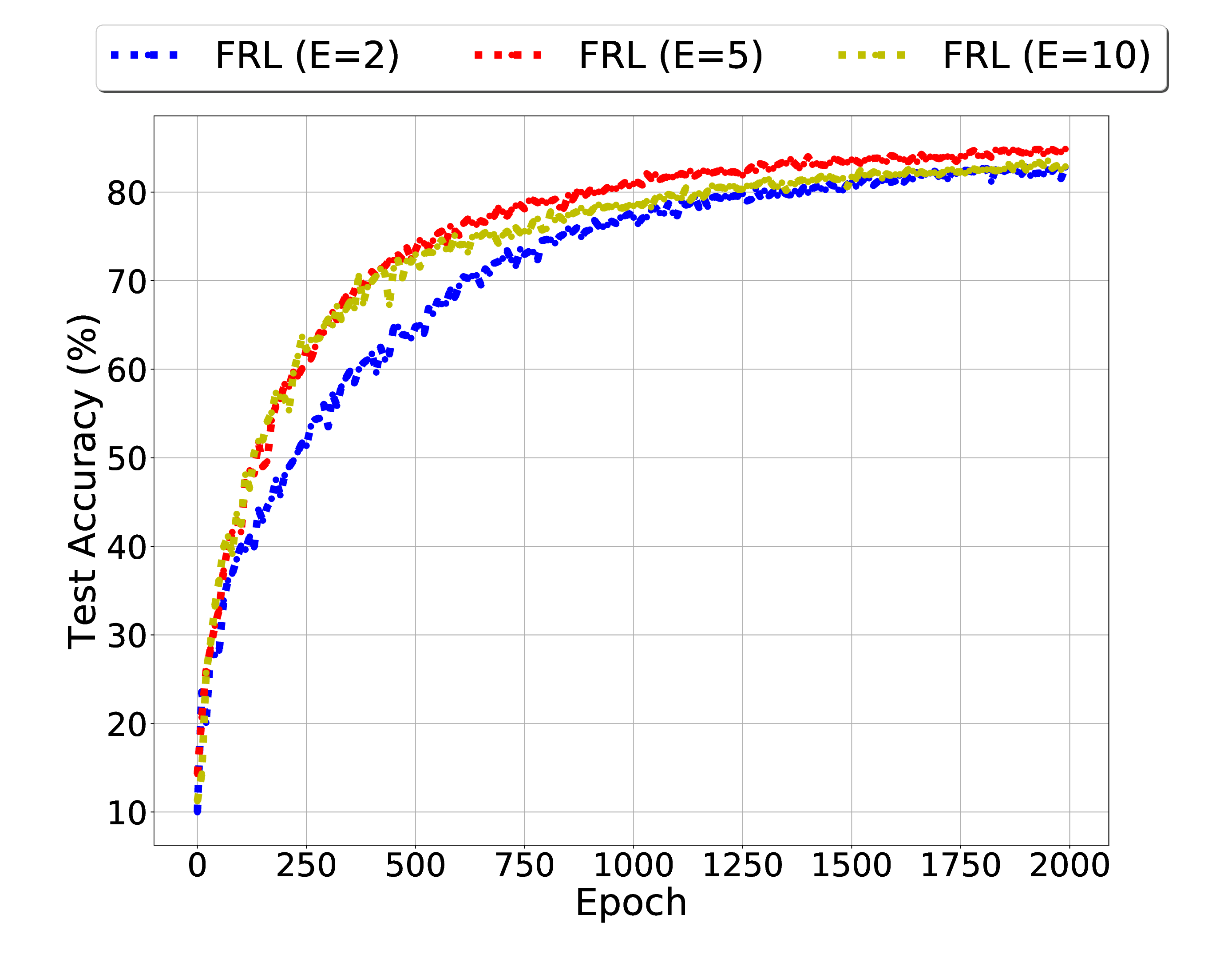}}
\end{subfigmatrix}
\begin{subfigmatrix}{2}
\centering
 \subfigure[CIFAR10 (Test Loss)]{
  \includegraphics[width=1\columnwidth]{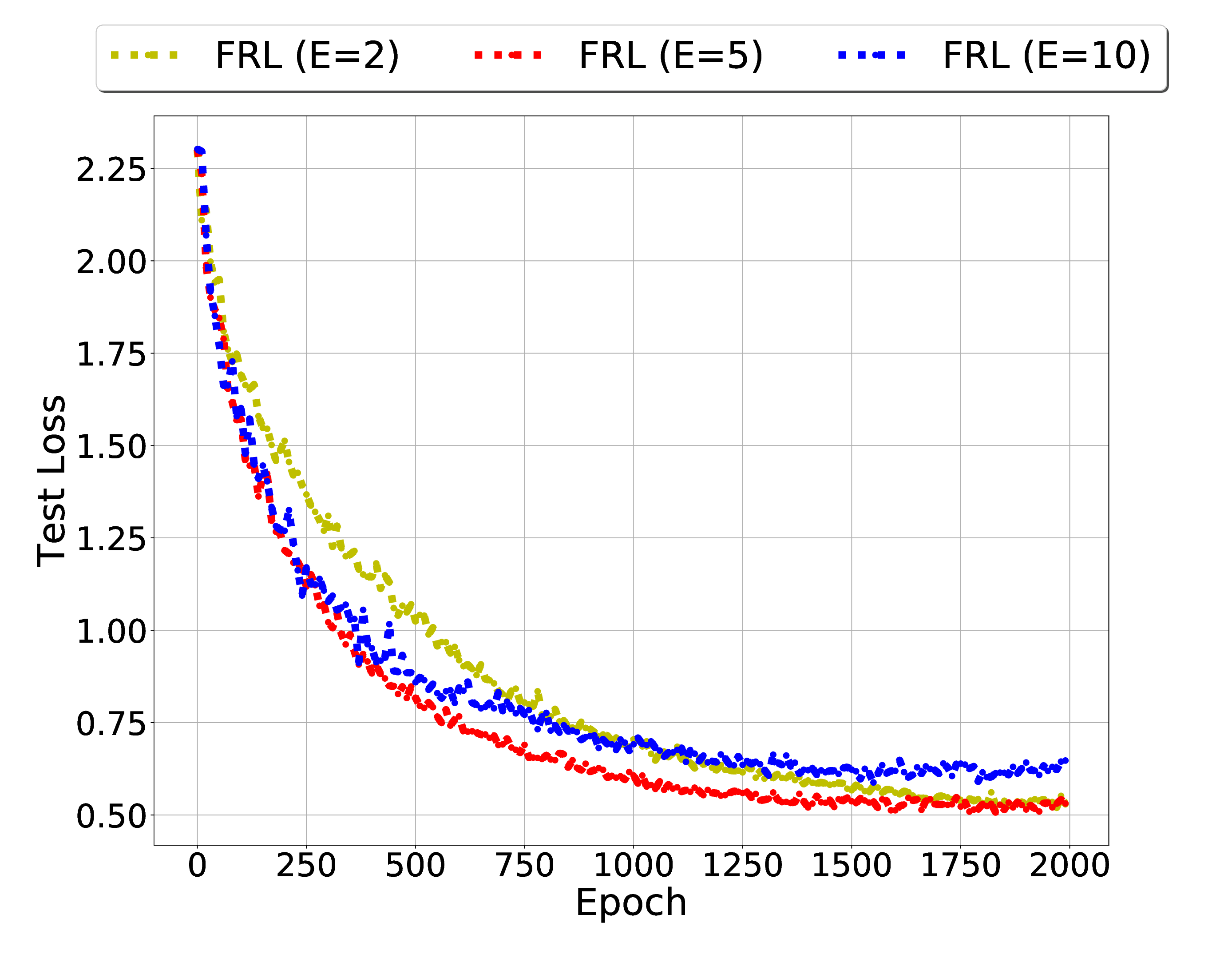}}
\end{subfigmatrix}
\vspace{-10pt}
\caption{Comparing performance of \Name{} for different local epochs. }
\label{fig:curve}
\vspace{20pt}
\end{figure}

Figure~\ref{fig:curve} is showing the learning curve of \Name{} for different numbers of local epochs for CIFAR10 experiment. The data is distributed non-iid using Dirichlet distribution. As we can see using 5 local epochs produces the best results. 

Table~\ref{tab:others} is showing the effect of other settings on performance of \Name{} trained on CIFAR10 distributed over 1000 clients using Dirichlet distribution. The \textbf{bold} shows the value we used in our experiments.

\begin{table}[H]
\caption{The effect of other settings on performance of \Name{} trained on CIFAR10 distributed over 1000 clients using Dirichlet distribution. The \textbf{bold} shows the value we used in our experiments. }\label{tab:others}
\centering
\setlength{\extrarowheight}{0.05cm}
\begin{tabular} {|c|c|c|c|}
  \hline
  Method & hyperparameter & value & \vtop{\hbox{\strut Test Accuracy }\hbox{\strut (10\% malicious)}} \\ \cline{1-4}
 \multirow{9}{*}{\Name{}} &  \multirow{3}{*}{ \vtop{\hbox{\strut  Number of }\hbox{\strut participants (n)}}} & 15  & 84.8 (11.3) \\ \cline{3-4}
 & & \textbf{25} &   \textbf{85.3 (11.3)} \\ \cline{3-4}
 & & 50 & 84.9 (11.2) \\ \cline{2-4}
 &  \multirow{3}{*}{local epochs (E)} & 2 & 82.2 (12.0) \\ \cline{3-4}
 & & \textbf{5}  & \textbf{85.3 (11.3)}  \\ \cline{3-4}
 & & 10  &  83.5 (11.9) \\ \cline{2-4}
 &  \multirow{3}{*}{Non-iid degree ($\beta$)} & \textbf{1} & \textbf{85.3 (11.3)} \\ \cline{3-4}
 & & 10 & 85.6 (11.1) \\ \cline{3-4}
 & & 100  & 85.6 (10.9) \\
\hline 
\end{tabular}
\end{table}

\end{document}